\documentclass[pdflatex,sn-mathphys]{sn-jnl}

\jyear{2023}%

\theoremstyle{thmstyleone}%
\theoremstyle{thmstyletwo}%

\theoremstyle{thmstylethree}%

\usepackage{graphics} 
\usepackage{epsfig} 
\usepackage{amsmath} 
\usepackage{subfigure}
\usepackage{multirow}
\usepackage{multicol} 
\usepackage{graphicx, epstopdf}
\usepackage{xspace,epsfig,url}
\usepackage{enumerate}
\usepackage{float}
\usepackage{epsf}
\usepackage{psfrag}
\usepackage{verbatim}
\usepackage[all]{xy}
\usepackage{color,soul}
\usepackage{bm}
\usepackage{comment}
\usepackage{cases}
\usepackage{hyperref}
\usepackage{physics}
\usepackage{amssymb}
\usepackage{pifont}
\usepackage{array} 

\begin{document}

\title[Shared-Control Teleoperation Paradigms on a Soft-Growing Robot Manipulator]{\bf Shared-Control Teleoperation Paradigms on a Soft-Growing Robot Manipulator}


\author*[1]{\fnm{Fabio} \sur{Stroppa}}\email{fabio.stroppa@khas.edu.tr}

\author[2]{\fnm{Mario} \sur{Selvaggio}}\email{mario.selvaggio@unina.it}

\author[3]{\fnm{Nathaniel} \sur{Agharese}}\email{agharese@stanford.edu}

\author[4]{\fnm{Ming} \sur{Luo}}\email{ming.luo@wsu.edu}

\author[5]{\fnm{Laura H.} \sur{Blumenschein}}\email{lhblumen@purdue.edu}

\author[6]{\fnm{Elliot W.} \sur{Hawkes}}\email{ewhawkes@ucsb.edu}

\author[3]{\fnm{Allison M.} \sur{Okamura}}\email{aokamura@stanford.edu}

\affil*[1]{\orgdiv{Computer Engineering Department}, \orgname{Kadir Has University}, \orgaddress{\city{\.Istanbul}, \postcode{34083}, \country{Turkey}}}

\affil[2]{\orgdiv{Electrical Engineering and Information Technology Department}, \orgname{University of Naples Federico II}, \orgaddress{\city{Naples}, \postcode{80125}, \country{Italy}}}

\affil[3]{\orgdiv{Mechanical Engineering Department}, \orgname{Stanford University}, \orgaddress{\city{Stanford}, \postcode{94305}, \state{CA}, \country{USA}}}

\affil[4]{\orgdiv{School of Mechanical and Materials Engineering}, \orgname{Washington State University}, \orgaddress{\city{Pullman}, \postcode{99163}, \state{WA}, \country{USA}}}

\affil[5]{\orgdiv{School of Mechanical Engineering}, \orgname{Purdue University}, \orgaddress{\city{West Lafayette}, \postcode{47907}, \state{IN}, \country{USA}}}

\affil[6]{\orgdiv{Mechanical Engineering Department}, \orgname{UC Santa Barbara}, \orgaddress{\city{Santa Barbara}, \postcode{93105}, \state{CA}, \country{USA}}}


\abstract{Semi-autonomous telerobotic systems allow both humans and robots to exploit their strengths while enabling personalized execution of a remote task.
For soft robots with kinematic structures dissimilar to those of human operators, it is unknown how the allocation of control between the human and the robot changes the performance.
This work presents a set of interaction paradigms between a human and a remote soft-growing robot manipulator, with demonstrations in both real and simulated scenarios. 
The soft robot can grow and retract by eversion and inversion of its tubular body, a property we exploit in the interaction paradigms. 
We implemented and tested six different human-robot interaction paradigms, with full teleoperation at one extreme and gradually adding autonomy to various aspects of the task execution. 
All paradigms are demonstrated by two experts and two naive operators.
Results show that humans and the soft robot manipulator can effectively split their control along different degrees of freedom while acting simultaneously to accomplish a task. 
In the simple pick-and-place task studied in this work, performance improves as the control is gradually given to the robot's autonomy, especially when the robot can correct certain human errors.
However, human engagement is maximized when the control over a task is at least partially shared. 
Finally, when the human operator is assisted by haptic guidance, which is computed based on soft robot tip position errors, we observed that the improvement in performance is dependent on the expertise of the human operator.
}

\keywords{Shared Control, Teleoperation, Human-Machine Interaction, Haptics, Soft Robotics}

\pacs[Type of Paper]{Category (3) 
}

\pacs[MSC Classification]{
Artificial intelligence for robotics (68T40), 
Automated systems in control theory (93C85), 
Computer science (68W99), 
Kinematics of mechanisms and robots (70B15), 
Control of mechanical systems (70Q05)}

\maketitle

\section{Introduction}

{R}{obots} in domestic environments, especially soft robots, have the potential to both autonomously assist humans and enable safe physical presence for remote operators~\cite{de2016long}. The spectrum of potential operation modalities in between full autonomy and direct human operation can provide robotic systems with a rich and useful set of perception and manipulation capabilities; however, its concrete realization also creates challenges for system designs and the development of their intuitive human-in-the-loop control~\cite{scalise2018balancing, schilling2016towards, CoadRetraction2020, el2018development}. This is especially true given the challenges inherent with soft robots, where humans may have advantages in understanding and controlling the underactuated robot state. 
Shared control denotes operation modalities of a robotic system that serve to balance inaccuracies of both human and artificial-intelligent agents, such that they can benefit from each one's abilities~\cite{mortl2012role, corrales2012cooperative}. Employing shared control becomes essential when humans are needed to address difficult-to-automate sub-tasks, yet some autonomy is needed to improve speed and/or performance and to reduce the physical/cognitive burden on human operators. 
For instance, most previous studies on shared-control telerobotics rely on human intelligence to solve low-level perceptual sub-tasks that are difficult for autonomous robots, allocating the sub-task control either to the human or to the robot control system, or partially to both~\cite{pitzer2011towards, kortenkamp2000adjustable, sellner2005user, dias2008sliding}. 

\begin{figure}[t!]
  \centering
	{\includegraphics[width=0.9\linewidth]{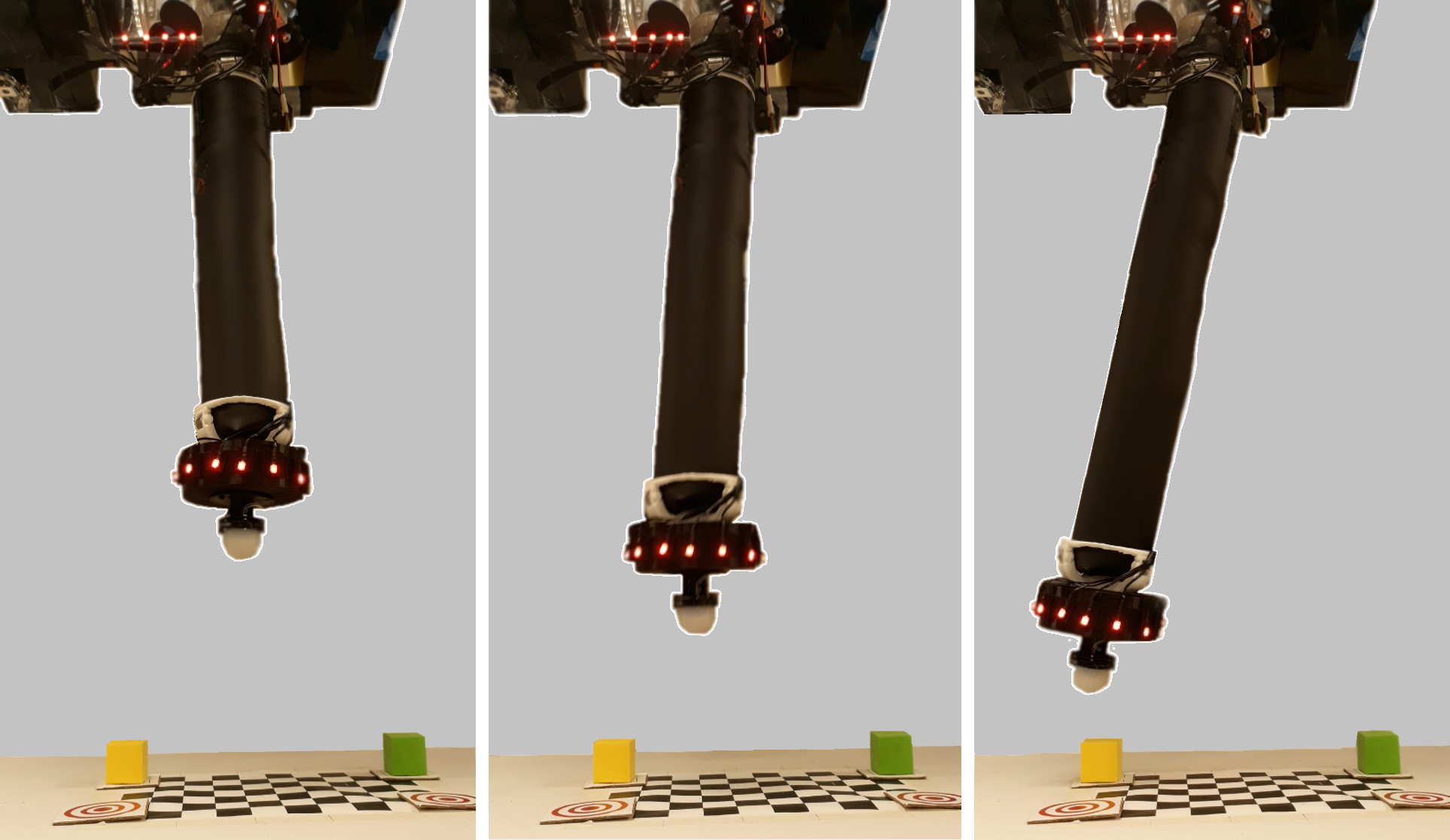}}\vspace{0.3cm}
    \caption{The soft-growing robot manipulator can extend from a base container by growing its body and steering it towards items in the workspace to perform pick-and-place tasks by means of a soft-magnetic gripper mounted at its tip.}
    \label{fig:robot_motion}
\end{figure}

In this work, we present the design and the demonstration of six shared-control paradigms that can be used by human operators to remotely interact with a robotic manipulator during a pick-and-place task. As a pilot investigation of novel shared-control paradigms, we deployed them in a straightforward scenario to understand the contributions of human and artificial agents. We were interested in scenarios in which the human perspective is distorted such that the robot could provide assistance. During remote teleoperation, humans operate a device that does not align with the representation of their body, differently from a scenario with wearable exoskeletons where the human perspective is unchanged. This reveals the extent to which autonomous features change the operator's experience and impact performance. We establish how humans can effectively interact with soft-growing robotic manipulators and which interaction paradigms (and specific aspects of those paradigms) are subjectively preferred. These developments can be extended to work in more complex scenarios in which the presence of the human is a requirement and the robot can either assist or be assisted by the human (e.g., exploration of cluttered or dangerous environments in which artificial perception is limited).

Besides sub-task allocation, the proposed shared-control paradigms feature different assistance levels (in the form of autonomy and/or haptic guidance) to help the operator throughout the task execution. We consider assistance in the form of (i) guidance force rendered on a haptic device, opportunely tuned based on the abilities of the operator, and (ii) a division of the robot's degrees of freedom (DoFs) between human and artificial/autonomous control. These forms of assistance are purposely designed to be suited when applied to a pick-and-place task and are independent of the employed robotic manipulator. 
We chose to demonstrate how these methods perform using a soft-growing robotic system (Fig.~\ref{fig:robot_motion}) with the aim of showing how they can be applied to this novel form robotic manipulator.

The use of shared-control teleoperation on a soft-growing robot manipulator has not been investigated yet to the best of our knowledge. Two of our proposed shared-control paradigms are novel and specifically designed for soft-growing manipulators, as they explicitly account for the shared actuation of unique DoFs. Unlike conventional rigid-body robots, the unique kinematics and compliance of \emph{soft-growing robots} have the potential to combine computational and embodied intelligence to facilitate manipulation in complex, unknown environments; and they also offer a safer and effective solution in environments with humans~\cite{rieffel2014growing, coad2019vine}. They imitate plant-like growth to change length and navigate environments, being able to grow and shrink through \emph{eversion} of thin-walled material at the tip~\cite{hawkes2017soft, sadeghi2017toward, tsukagoshi2011tip, dehghani2017design}. Their high portability makes them great candidates for applications where rigid robots cannot be employed due to the soft-growing robot properties of low inertia, low weight, and the ability to be fully stowed in small containers when retracted. They have been employed to explore an archaeological site~\cite{coad2019vine}, deploy reconfigurable structures~\cite{blumenschein2018tip}, navigate through coral reefs~\cite{luong2019eversion}, deliver tools~\cite{hawkes2017soft}, and burrow through granular media~\cite{naclerio2018soft}. However, their dynamic properties and non-anthropomorphic motion capabilities demand novel remote control paradigms for such scenarios~\cite{el2018development, stroppa2020human}.

\section{Background on Shared Control}
\label{sec:shared_control_background}

The term \emph{shared control} generally refers to a scenario in which both human and robotic agents work (sometimes remotely) together to accomplish a common task. Broadly it is the spectrum of possible interactions between humans and robots, from robots having full autonomy to none at all~\cite{goodrich2008human}.
The development of shared-control techniques must account for several aspects such as the type of interaction, the type of task being accomplished, the forms of user feedback, etc. 
This results in the definition of semi-autonomous control strategies with different and possibly varying levels of autonomy and user feedback~\cite{SelvaggioRAL2021}. 
Full autonomy still poses a problem for robotic systems when dealing with unknown or complex tasks in unstructured and uncertain scenarios~\cite{Yangeaam8638}; on the other hand, shared control has been demonstrated to improve the task performance without placing the entire burden on the human operator~\cite{kanda2017human}. Several works about shared control focus on the extent of human intervention in the control of artificial systems, conceding a certain amount of responsibility to each agent in the scene so as to split the control burden among all the participants~\cite{schilling2019shared}. 
The extent of human intervention, and thus robot autonomy, has been usually classified into discrete levels~\cite{dias2008sliding, bruemmer2002dynamic, kortenkamp2000adjustable}, with fewer studies considering a continuous domain~\cite{desai2005blending, anderson2010semi}. Commonly, shared-control techniques aim to fully or partially replace a function, like identifying objects in cluttered environments~\cite{pitzer2011towards}, while others start from an autonomous robot autonomously and give control to the user only in difficult situations~\cite{kortenkamp2000adjustable, sellner2005user, dias2008sliding}. 

In a manipulation scenario, shared control has been used to decrease the cognitive and physical workload of the human operator, with the robot autonomously carrying out a set of secondary tasks~\cite{selvaggio2019passive}. Some works suggest that the robot can autonomously perform grasp when close to the goal~\cite{kofman2005teleoperation}, hold it over long time periods~\cite{griffin2005feedback}, or switch the control at some trigger~\cite{li2003recognition, kofman2005teleoperation, kragic2005human, shen2004collaborative}.
Sharing the control can also be seen as a means of collaborative interaction~\cite{schilling2016towards}. It can be interpreted as \textit{assistance}, not only from the human to the robot but also vice versa. The latter is advantageous when the teleoperation is hindered by the inadequacies of the input control commands or by poor perception (e.g., remote teleoperation in which the point of view of the operator is not directly over the workspace). The robot may assist the user in accomplishing the desired task, making the teleoperation easier~\cite{dragan2013policy}. Therefore, the task of the robot becomes modulating the output rather than simply executing the operator's input.
Some studies assist the operator by predicting their intent while selecting among different targets~\cite{dragan2013policy, javdani2015shared}, while others exploit haptic feedback/guidance techniques while moving toward a specific target~\cite{crandall2002characterizing, aarno2005adaptive, selvaggio2016enhancing, selvaggio2018haptic, selvaggio2019haptic}.
%
%
These types of haptic assistance enforce constraints on the operator's position by applying controlled forces that are a function of the evolving behavior of the task or system~\cite{o2006shared}. Alternatively, paradigms such as \emph{assist-as-needed} constrain the operators only when their behavior is in conflict with a known task~\cite{squeri2011adaptive, guidali2011online, stroppa2017online}.

\begin{figure}[h!]
\centering
	\subfigure[\protect\url{}\label{fig:real}]%
{\includegraphics[height=4cm]{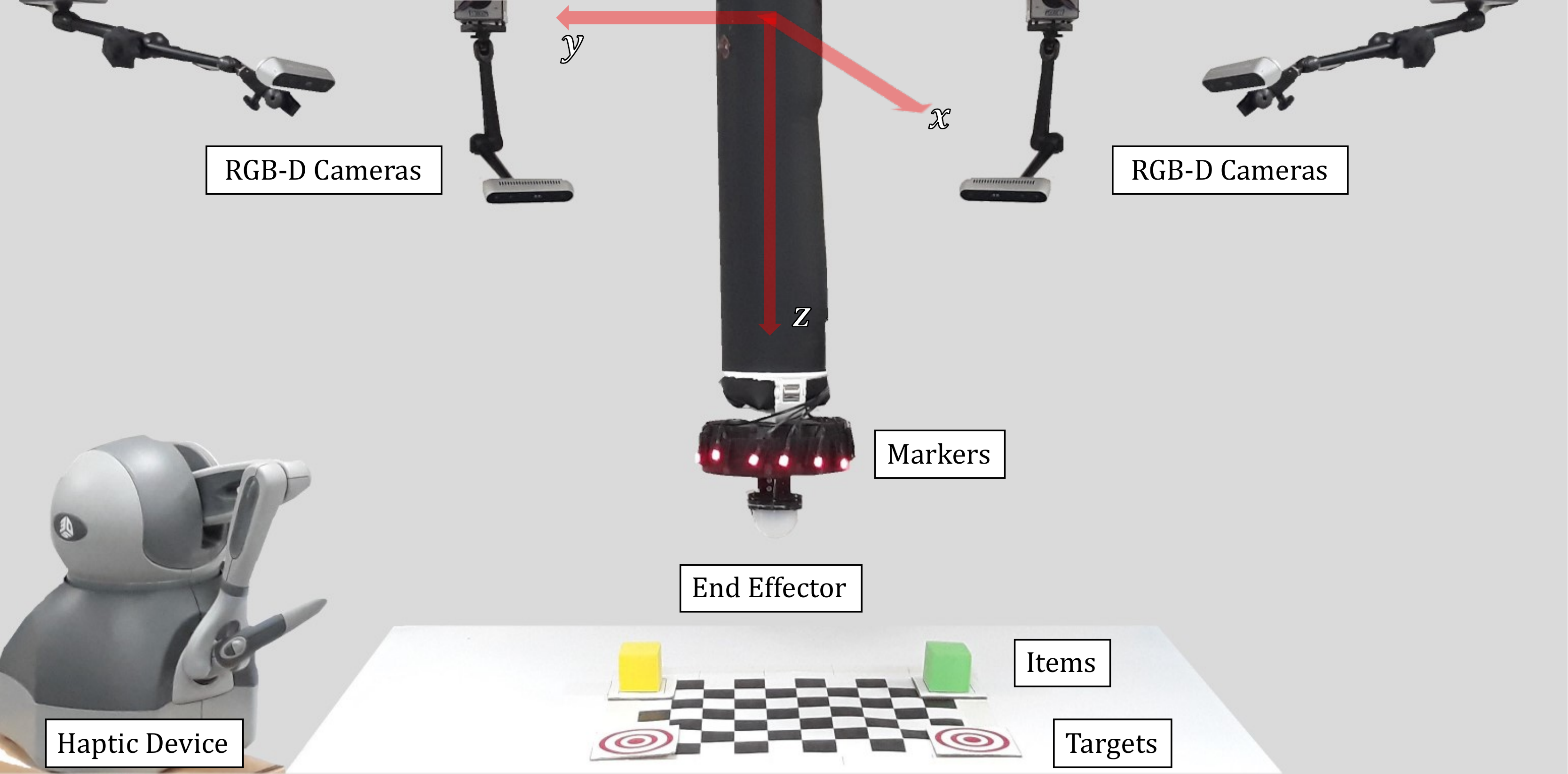}}\qquad
\subfigure[\protect\url{}\label{fig:simulation}]%
{\includegraphics[height=4cm]{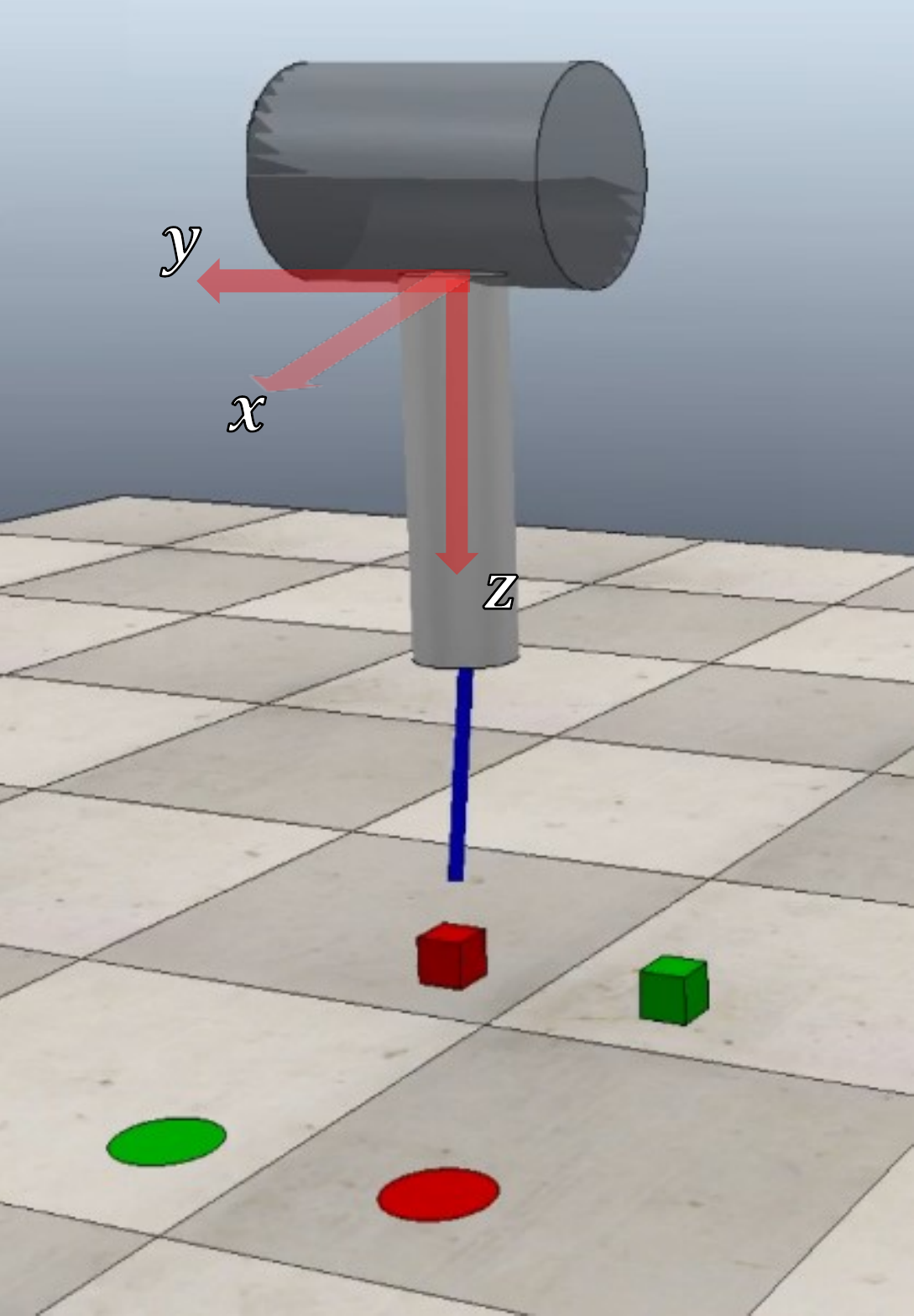}}
	\subfigure[\protect\url{}\label{fig:scheme}]%
	{\includegraphics[width=\textwidth]{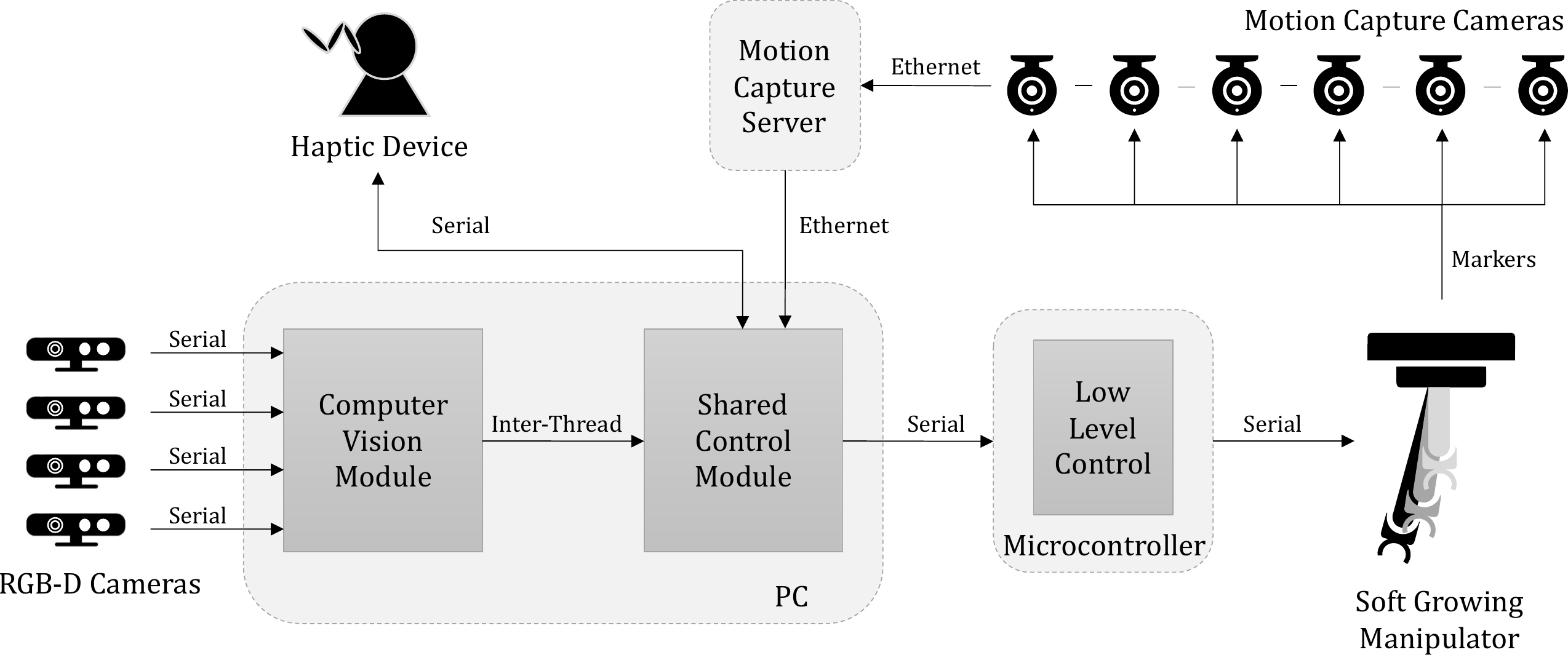}}\vspace{0.3cm}
	\caption{(a) Picture of the real soft-growing robotic manipulation system. A coordinate system is attached to the robot base: the $z$ axis is oriented along the gravity direction, while the $x$ and $y$ define the axes around which the robot steers at the base. As such, the soft-growing manipulator has 3 degrees of freedom, one associated with growth and two associated with steering. The items and targets associated with the pick-and-place task are laying on the table at the bottom.
	(b) Picture of the simulated soft-growing manipulator as well as items and targets. The robot and its reference system are oriented as in the real system (a). The blue vector represents the assistive force displayed to the user during assisted shared-control operations.
	(c) Block scheme of the overall system. The Shared Control and the Computer Vision modules run on a personal computer (PC), while Low-Level Control runs on a microcontroller connected to the manipulator. A Motion Capture system tracks the position of the manipulator while a set of RGB-D Cameras tracks the items in the environment. The human operator interfaces with the manipulator via the Haptic Device. Each connection between components indicates the type of communication (e.g., through an Ethernet cable) and whether it is one-way or two-way (single-ended or double-ended arrows, respectively).}\label{fig:setup}
\end{figure}

\section{System Description}
\label{sec:system}
This section describes the compliant robotic system used to study the proposed soft robot shared-control teleoperation paradigms for a pick-and-place manipulation task. The main components of the system are the soft-growing manipulator, both in real (Fig.~\ref{fig:real}) and simulated scenarios (Fig.~\ref{fig:simulation}), a kinesthetic haptic device, used to collect inputs and display haptic guidance to the operator, and two tracking systems based on motion capture and computer vision, that retrieve the robotic tip position and the items/targets of the pick-and-place task (in Cartesian pose), respectively. Figure~\ref{fig:scheme} shows a block scheme of the system highlighting the interconnections among its modules. 

The pick-and-place task consists of reaching, grasping, and placing two foam cubes in the corresponding target locations of the workspace. In the following, we will refer to the main components of the task as:
\begin{itemize}
    \item \textit{item}: the generic foam cube to be grasped and relocated;
    \item \textit{target}: the place where the foam cube should be relocated; and 
    \item \textit{goal}: the current objective, which refers to a specific item when the gripper does not hold any, or to a target when the gripper is already holding an item. 
\end{itemize}



\subsection{Soft-Growing Manipulator}
\label{sec:robot_and_gripper}

The robotic manipulation system consists of a soft-growing manipulator that can grow, retract, and steer in three dimensions (see Fig.~\ref{fig:robot_motion})~\cite{coad2019vine}.  We used the same manipulator for our previous works on teleoperation~\cite{stroppa2020human}. Fig.~\ref{fig:real} shows a close-up of the manipulator alongside its reference frame: markers are placed on the end-effector to retrieve its Cartesian position. 

\begin{figure}[h!]
	\centering
	{\includegraphics[width=\textwidth]{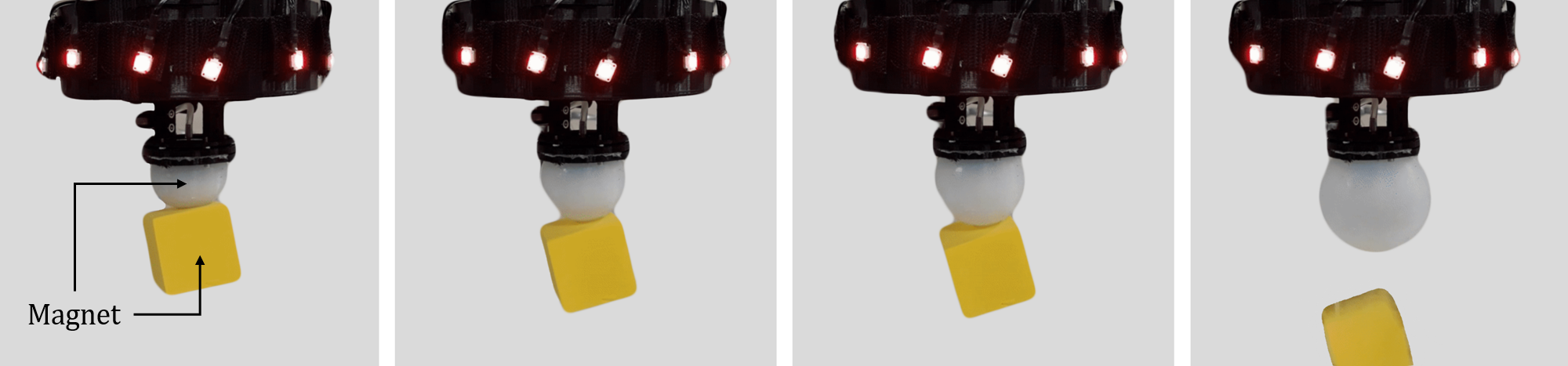}}\vspace{0.3cm}
	\caption{Sequence of pictures showing the soft-magnetic gripper inflating and releasing an item. The gripper and the item both have a magnet inside, with opposite polarities. When the gripper is inflated, the distance between the two magnets increases until the attraction force is lower than the item's gravity and the item is dropped.}
	\label{fig:gripper}
\end{figure}

The manipulator can carry a payload thanks to the soft-magnetic gripper attached to the end effector.
The gripper can magnetically attract objects without requiring the precise positioning of the end-effector. We placed cylindrical magnets inside the gripper's silicone cover and inside the target items (with opposite polarity) so that the gripper can attract them when sufficiently close. To release items, we inflate the soft gripper such that the silicone rubber pushes the grasped item away from the magnet, increasing the distance between the two magnets until the attraction force is low enough to drop the item (Fig.~\ref{fig:gripper}).

\subsection{Haptic Device(s)}
\label{sec:haptic}
We used a Phantom Omni (now marketed as Geomagic Touch) to teleoperate the robot's end-effector and perform the task in the real scenario (as shown in Fig.~\ref{fig:real}), and a Novint Falcon (Novint Technologies, Inc.) in the simulated one. These kinesthetic haptic devices provide force rendering up to $7$~N along their linear directions. Linear displacements are used to command the manipulator's tip position in the real scenario and its tip velocity in simulation such that the haptic device's workspace is scaled to the position or velocity workspace. 
The haptic device is equipped with two buttons, which are used to inflate/deflate the gripper. The haptic devices are also used to render haptic forces to the user (see Sec.~\ref{sec:shared_autonomy_paradigms}).

\subsection{Tracking Modules}
\label{sec:tracking_modules}
We used two different systems to track the manipulator's end-effector (see Sec.~\ref{sec:mocap}) and the items to be manipulated (see Sec.~\ref{sec:cvmod}), based on motion capture and computer vision, respectively. Both tracking modules are transformed and expressed in the reference system depicted in Fig.~\ref{fig:real} to align with the kinematics and control of the robot. 

In our study, we used a controlled environment to focus on the human-machine interaction; however, the implemented motion capture system would not be available in all real-world scenarios. Methods are being developed to localize soft robots' tip position in the environment, such as using cameras placed in an eye-in-hand configuration~\cite{hawkes2017soft, greer2019soft, CoadThesis2021} or shape sensors along the length~\cite{gruebele2020distributed}.

\subsubsection{Robot Tracking}
\label{sec:mocap}
We used the PhaseSpace Impulse X2E to track the manipulator's end-effector. This motion capture system allows real-time tracking at $270$~Hz. The cameras are managed by an external server, which communicates with the main personal computer (PC) via ethernet connection (Fig.~\ref{fig:scheme}). As shown in Fig.~\ref{fig:real}, eight markers are placed around the circumference of the gripper mount at the manipulator's end-effector and tracked by six cameras located on the ceiling (cameras are not shown in the picture). The position of the end-effector is given by the markers' centroid.

\subsubsection{Item Tracking}
\label{sec:cvmod}

Four RGB-D cameras (RealSense D415) surround the workspace and are used for item tracking. 
The data captured by the cameras (2D image and depth map for 3D reconstruction) is processed by a Computer Vision Module installed on the main PC (Fig.~\ref{fig:scheme}). This module aligns the data retrieved by different cameras through automatic calibration of their extrinsic parameters (camera resectioning), such that each camera shares the same reference system. The calibration is independent of the cameras' locations, as they face the same reference system -- a physical chessboard placed in the center of the workspace as shown in Fig.~\ref{fig:real}. 
Different items are identified by their color. 

Although this method could potentially run in real time
, we simplified the system by measuring the positions of the items only once before a trial and storing them as fixed values (when the manipulator approaches an item, it occludes the scene and the items cannot be tracked; this hinders the opportunity to identify whether the gripper is currently holding an item for full automation purposes.)
Similarly, the static targets are identified only once, by positioning an item on it and storing that position as a target. Future works could expand this system to integrate state-of-the-art tracking and handle moving targets; for the purpose of our shared-control studies, we implemented the basic functionality required for evaluating the spectrum of teleoperation and quasi-autonomous operations.



\subsection{Simulated Scenario}
\label{sec:simulated}

A soft-growing robot simulator was set up in CoppeliaSim~\cite{rohmer2013v} to develop and preliminary test our shared-control paradigms. 
The virtual environment is a digital twin of the real robotic system with items and targets placed in the same relative positions with respect to the manipulator (see Fig.~\ref{fig:simulation}).
The manipulator was digitally modeled as a cylindrical body that can scale its length along its main axis to simulate eversion. 
%
%
%
When assistive paradigms are used, the haptic force rendered through the haptic devices is visually displayed to the user as a blue line segment at the robot tip as shown in Fig.~\ref{fig:simulation}.
%

The purpose of the simulation was to systematically tune the control parameters for the shared control paradigms featuring haptic assistance and demonstrate their performance in ideal conditions (see Sec.~\ref{sec:params}). Besides the unconventional kinematics of soft-growing robots, their motion fluidity is affected by several mechanical factors (actuation dynamics, friction, and external disturbances), which may have an influence on the effectiveness of shared control paradigms. Thus, we conducted a parameter-tuning simulation to establish the main effects and the interactions of the shared-control paradigm parameters in ideal simulated conditions. Then, we compared the simulated and the real system to demonstrate how these mechanical factors affect the effectiveness of our shared-control paradigms.


\section{Shared-control Teleoperation Paradigms}
\label{sec:shared_autonomy_paradigms}

In this section, we describe the six proposed shared-control teleoperation paradigms used to investigate the level of autonomy needed for a human to accomplish a pick-and-place task with a soft-growing robot manipulator.
We included the case in which the control is not shared as a baseline for the comparison (i.e., direct teleoperation, in which the human is in full control of the system, and then built from there by gradually adding autonomy until the operator only has control over the gripper operation. 
The order in which the paradigms are presented follows increments in autonomy, as the control of more DoFs shifts from the human to the robot. 
Note that most of these paradigms are generic and can be applied to any kind of robotic manipulator that can be commanded in Cartesian space; however, two of them (see Sec.~\ref{sec:msae} and~\ref{sec:asme}) are specific for soft robots in that they explicitly exploit eversion. 

We will use the following terms to denote quantities measured with respect to the robot's base frame:
\begin{itemize}
    \item $ee = \left\langle x , y , z\right\rangle \in \mathbb{R}^3$ is the current position of the manipulator's end-effector, coincident with the tip of the gripper;
    \item $d = \left\langle x , y , z\right\rangle \in \mathbb{R}^3$ is the desired position of the manipulator's end-effector (i.e., where the manipulator is commanded to go), coincident with the tip of the gripper;
    \item $c = \left\langle x , y , z\right\rangle \in \mathbb{R}^3$ is the command given by the haptic device that is to be mapped to $d$; and
    \item $g = \left\langle x , y , z\right\rangle \in \mathbb{R}^3$ is the position of the current goal.
\end{itemize}
The term $d$, that is the desired position of the manipulator's end-effector, is set to either $c$ or $g$, or a combination of them, based on the type of shared control modality. We indicate with $m=\norm{g-ee} \in \mathbb{R}_{\geq 0}$ the Euclidean distance between the goal and the end-effector at a current time and with $\dot{m} \in \mathbb{R}$ its time derivative, which is used to evaluate the motion performed by the robot.
Inverse kinematics is used to compute growth and steering;
whereas 
low-level control is used to bring $ee$ towards $d$.

 \begin{figure}[t]
  \centering
	{\includegraphics[width=10cm]{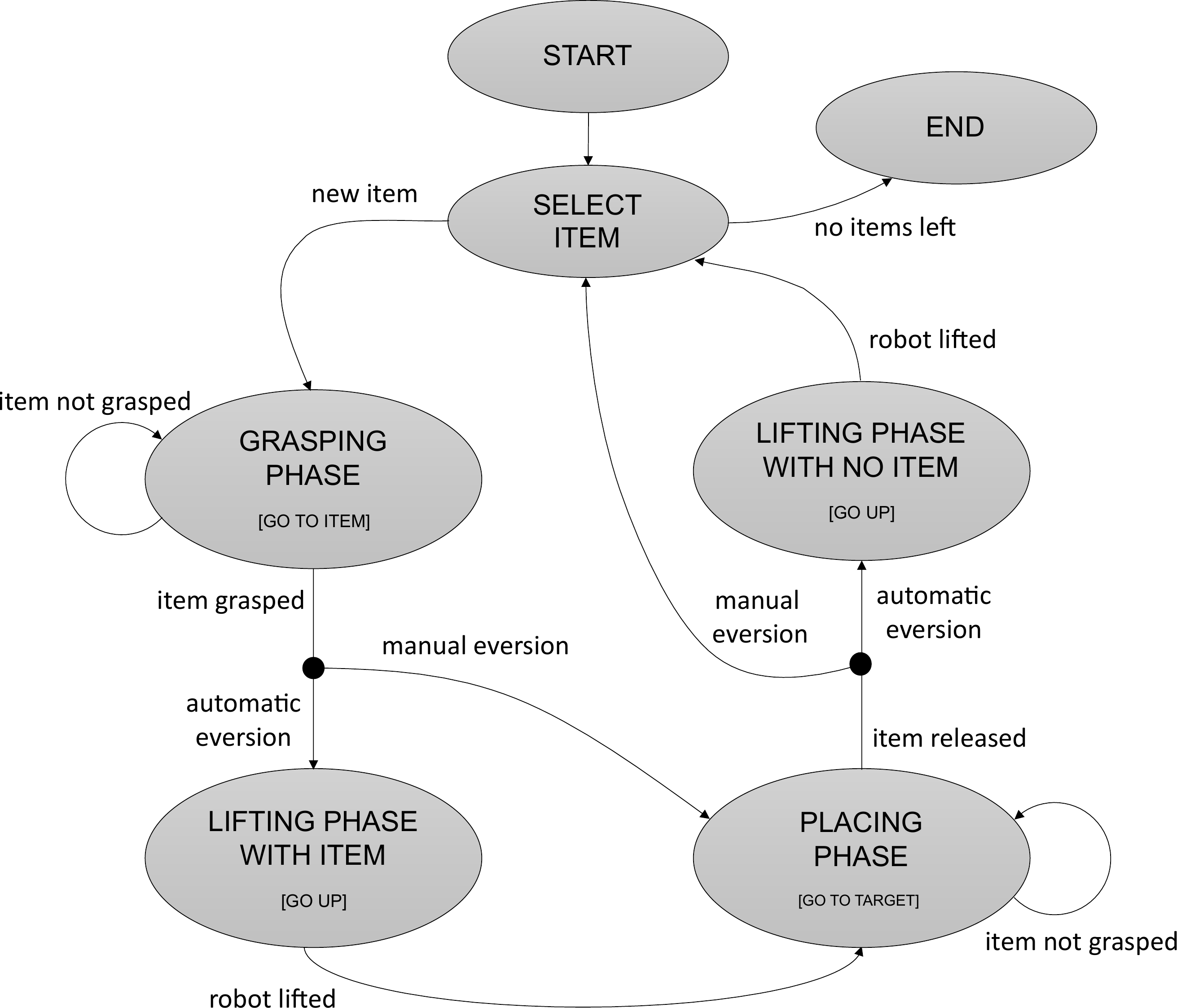}} \vspace{0.3cm}
    \caption{State machine for the Goal Selection algorithm, used to monitor the evolution of the pick-and-place task and select the appropriate goal. The current goal can be either an object to grasp (item) or a desired position of placing (target), depending on whether the robot is currently grasping an item with the gripper. Additionally, if the eversion is automatically performed, the system will slightly lift the robot before moving on to the next goal, preventing the gripper to be dragged onto the ground. }
    \label{fig:state_machine_GS}
\end{figure}

During the execution of a task, $g$ changes based on whether the current goal is the position of an \textit{item} or a \textit{target}. 
To handle this procedure, we developed a \textit{Goal Selection} algorithm performing the following steps:
(i) select the item to be grasped (grasping phase);
(ii) select the target where the current item will be placed (placing phase); and 
(iii) repeat until all the items are correctly placed in their respective target position (the order in which the items are selected is beyond the purpose of the algorithm.) 
The algorithm is implemented as a state machine, graphically represented in Fig.~\ref{fig:state_machine_GS}.
Besides providing a set of locations that the manipulator is asked to reach, this algorithm allows the system to monitor the progress of the task.
A further behavior enabled by the algorithm occurs when a goal is reached, and thus the position of the end-effector is supposedly near the ground: in such a case, it is desirable to lift the robot before moving towards the next goal, preventing problems such as dragging the tip over the ground, or hitting other items on the way.
To handle this, the state machine introduces an intermediate goal: a proxy set above the current end-effector position away from the ground, resulting in the robot being lifted before switching to the next goal (lifting phase). Note that this intermediate goal is generated only if the growth DoF is automated (as will be described in Sec.~\ref{sec:asme} and~\ref{sec:mostly_autonomy}), because this is a  location in the space that the human operator would not visualize in a real environment.
The features of the six shared-control teleoperation paradigms are preliminary summarized in Table~\ref{tab:paradigmsSummary}. 

\begin{table}[h]
\tiny
\centering
\caption{Feature summary of the shared-control teleoperation paradigms}
\label{tab:paradigmsSummary}
\begin{center}
 \renewcommand\arraystretch{1.4}
{
 \begin{tabular}{p{1.8cm}| c c c c c }
 \hline
& Steering & Eversion & \begin{tabular}[c]{@{}c@{}}Grasping operation \\ (Gripper open/close)\end{tabular} & \begin{tabular}[c]{@{}c@{}}Grasping detection\\ (Item grasped yes/no)\end{tabular} & \begin{tabular}[c]{@{}c@{}}Haptic \\ Guidance\end{tabular} \\ \hline
\textit{Full Teleoperation} & manual & manual & manual & manual & no \\ 
\textit{AAN} & manual & manual & manual & manual & yes \\ 
\textit{Fixed Assistance} & manual & manual & manual & manual & yes \\ 
\textit{MSAE} & manual & automatic & manual & manual & no \\ 
\textit{ASME} & automatic & manual & manual & manual & no \\ 
\textit{Mostly Autonomy} & automatic & automatic & manual & manual & no \\ 
\end{tabular} }
\end{center}
\end{table}

Due to the grasping system described in Sec.~\ref{sec:robot_and_gripper}, 
we used the following assumption: once an item is grasped, it will only be released on a target. If the item is erroneously dropped anywhere else, the trial is discarded; although due to the nature of the soft-magnetic gripper, this never occurred during our trials. 
This assumption overcomes a practical limitation of our system: we have no means to track whether an item is being currently grasped. Firstly, item tracking is not performed in real time, but only at the beginning of each trial, therefore the positions of items and targets do not change over time. Secondly, since the soft-magnetic gripper may require variable time to release an item, we cannot exactly estimate when this happens without having a sensor in the gripper - which we did not include.
Thanks to the aforementioned assumption, we decided that in our implementation the human is always in charge of controlling the grasping operation (whether the gripper should be `opened' or `closed') and of performing grasp detection (whether the gripper is currently holding an item or not).
The operator can trigger the opening/closing of the gripper by pressing a button on the haptic device and indicates to the system that the item has been grasped/released by pressing a second button on the haptic device.

\subsection{Full Teleoperation}
\label{sec:full_teleoperation}

In this paradigm, the human operator has full control over the manipulator, in the sense that all the Cartesian displacements of the end-effector are commanded through the haptic device commands $c$. 
Thus, this paradigm is realized by setting: 

\begin{equation}\label{eq:fullTeleoperation}
    \begin{aligned}
        d = c
    \end{aligned}
\end{equation}
This paradigm is considered a baseline on which the other paradigms are built, but some could also consider it as shared control in that closed-loop control is the contribution of the robotic system.

\subsection{Assist-as-Needed (AAN)}
\label{sec:aan}

In this paradigm, the human operator still has full control over the manipulator while occasionally being assisted by a force rendered by the haptic device. The Assist-as-Needed paradigm is implemented such that the system monitors the movements of the operator, provides assistance according to this: if the operator is correctly moving toward the goal, then no assistance is needed, and any force previously generated is gradually reduced. Alternatively, if the operator is performing poorly (e.g., not moving or going in the opposite target direction), then the algorithm generates a force that increases over time while the performance is not improving. With this mechanism, sufficient force to overcome physical or perceptual impairment is provided to the operator. 

For this paradigm, we define $\vec{f} \in \mathbb{R}^3$ as the haptic guidance force pushing towards the goal, expressed in N, whose value cannot exceed $f_{MAX}$.
The direction (unit vector) of the assisting force $\hat{f} \in \mathbb{R}^3$ goes from the end-effector position to the goal. Its magnitude $\abs{f}$ depends on $k$, a (variable) parameter denoting the stiffness modulating the force using Hooke's Law, whose value cannot exceed $k_{MAX}$ (both force and stiffness are constrained). The parameter $k$ is expressed in N$/$m, but it is ultimately dependent on the base unit selected for $ee$ and $g$. The following equations are involved in the calculation of the haptic guidance force $\vec{f}$:
\begin{equation}\label{eq:hookslaw_1}
    \begin{aligned}
        \abs{f} &= \begin{cases}
            k\cdot m \quad \text{if} \quad <f_{MAX}\\
            f_{MAX} \qquad \quad \text{otherwise}
        \end{cases}\\
        \hat{f} & = \frac{g-ee}{m} \\
        \vec{f} & = \abs{f}\cdot\hat{f}
    \end{aligned}
\end{equation}

\begin{figure}[t!]
  \centering
	{\includegraphics[width=10cm]{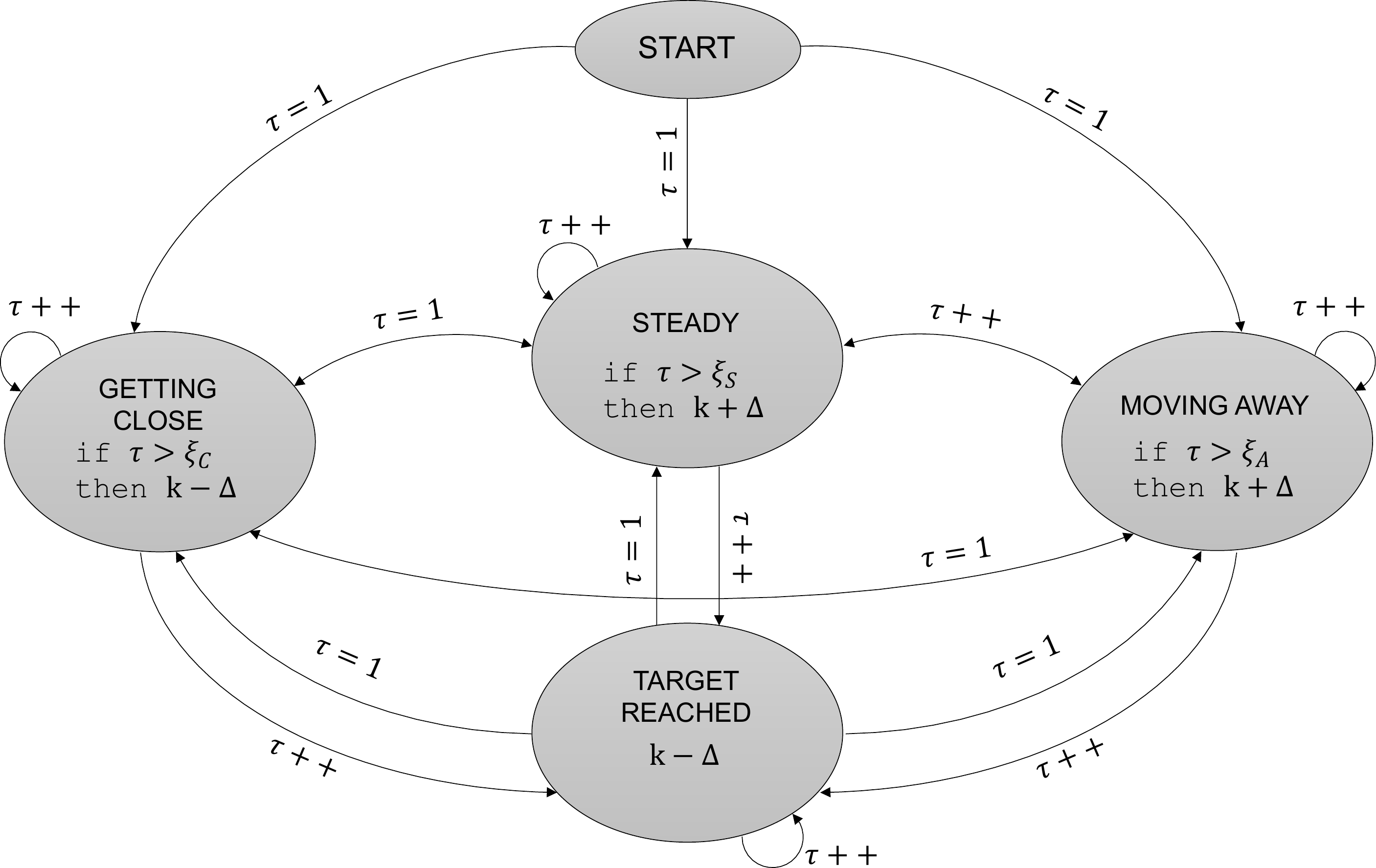}}\vspace{0.3cm}
    \caption{State machine used to evaluate the stiffness $k$ in the Assist-as-Needed shared-control teleoperation paradigm. The evaluation is based on the time elapsed in each state, monitored by the counter $\tau$. Each transition between states will either increment or reset $\tau$, and the value of $k$ will change only when $\tau$ is greater than the limit $\xi$ - e. g. after a certain amount of time spent in a certain state. The parameter $\xi$ may have a single value or a different value for each state.}
    \label{fig:state_machine}
\end{figure}

The procedure to modulate the stiffness parameter $k$ according to the task performance is implemented as a state machine (see Fig.~\ref{fig:state_machine}). Besides the starting state, there are four possible states in which the robot can be: (i) the operator has reached the goal ($m = 0$); (ii) the operator is steady ($\dot{m} = 0$); (iii) the operator is moving away from the goal ($\dot{m} > 0$); and
(iv) the operator is getting closer to the goal ($\dot{m} < 0$). This strategy increases the value of $k$ if the operator is steady or moving away from the goal, meaning that the robot should push in the direction of the goal as the operator is poorly teleoperating. On the contrary, the value of $k$ will decrease if the operator is getting closer to the goal, as the robot is already moving in the right direction. In any case, between two consecutive iterations, $k$ will change of at most an amount $\Delta$, a fixed positive value expressed in N/m, which will be added or subtracted to $k$ based on the state.

Since we want the operator to be in charge of the teleoperation, the logic is to wait for the operator to perform an action. This is implemented by the parameters $\xi$, a reaction time before the algorithm modifies the output, and $\tau$, a counter to watch the reaction time, initialized whenever the state changes, and incremented by one at each iteration as long as the system remains the same state.
Both the parameters are expressed in a number of iterations, which can be transformed to a time base unit when multiplied by the algorithm's sample rate.
The reaction time could be a single parameter, or we could use a different value for each state. This is useful when we want to manage various situations differently: e.g., the algorithm will wait for a certain amount of time $\xi_S$ before beginning to modify $k$ when the end-effector is not moving, and a different amount of time $\xi_C$ when the robot is moving towards the goal, etc.

The equation below shows the paradigm formulation and how the derivative of $k$ changes based on the slope between two consecutive distances from the goal:

\begin{equation}
    \begin{aligned}\label{eq:Assist-as-Needed}
        d &= c\\
        \dot{k} &= \begin{cases}
            -\Delta \quad \text{if} \quad m = 0  \enskip \land \enskip  k > 0 \\
            \hspace{0.26cm} \Delta  \quad \text{if} \quad \dot{m} \geq 0 \enskip \land \enskip k < k_{MAX} \enskip \land \enskip \tau < \xi\\
            -\Delta \quad \text{if} \quad \dot{m} < 0  \enskip \land \enskip  k > 0 \enskip \land \enskip  \tau < \xi\\
            \hspace{0.28cm} 0  \quad \text{otherwise}
        \end{cases}
    \end{aligned}
\end{equation}


\subsection{Fixed Assistance}
\label{sec:fixed_assistance}

In this paradigm, the human operator still controls the whole manipulator DoFs through the haptic device while being constantly assisted by $\vec{f} \in \mathbb{R}^3$, a force pushing towards the goal, expressed in N, whose value cannot exceed $f_{MAX}$. This is in contrast to the Assist-as-Needed paradigm, which only occasionally assists the human operator.
The direction of the assisting force $\hat{f}$ goes from the position of the end-effector towards the goal; whereas its magnitude $\abs{f}$ is given by Hooke's Law as shown in~\eqref{eq:hookslaw_2}, in which: $k$ is a fixed stiffness expressed in N$/$m, but it is ultimately dependent on the base unit selected for $ee$ and $g$; and $b$ is a fixed damper expressed in N$\cdot$s$/$m, multiplied to the velocity of the end-effector $\dot{ee}$ (this velocity can refer to either the haptic device or the robot end-effector).

\begin{equation}\label{eq:hookslaw_2}
    \begin{aligned}
        \abs{f} &= \begin{cases}
            k\cdot m  + b \cdot \dot{ee} \quad \text{if} \quad <f_{MAX}\\
            f_{MAX} \qquad \qquad \qquad \text{otherwise}
        \end{cases}\\
        \hat{f} & = \frac{g - ee}{m}\\
        \vec{f} & = \abs{f}\cdot\hat{f}
    \end{aligned}
\end{equation}

The paradigm formulation is thus
\begin{equation}\label{eq:fixed_ass}
    \begin{aligned}
        d &= c \\
        \dot{k} &= 0\\
        \dot{b} &= 0\\
    \end{aligned}
\end{equation}
To avoid sudden changes in the assisting force, a moving average filter is applied to $\abs{f}$ over 50 samples at 100~Hz.


\subsection{Manual Steering, Autonomous Eversion (MSAE)}
\label{sec:msae}

In this paradigm, the operator only controls the steering DoFs ($x$ and $y$ coordinates of the reference frame), while the robot is in charge of everting towards the goal ($z$ coordinate of the reference frame). The paradigm formulation is:

\begin{equation}\label{eq:MSAE}
    \begin{aligned}
        d = \left\langle c_x,c_y,g_z\right\rangle
    \end{aligned}
\end{equation}

\subsection{Autonomous Steering, Manual Eversion (ASME)}
\label{sec:asme}

In this paradigm, the operator only controls the eversion DoF ($z$ coordinate of the reference frame), while the robot is in charge of steering towards the goal ($x$ and $y$ coordinates of the reference frame). The paradigm formulation is:

\begin{equation}\label{eq:ASME}
    \begin{aligned}
        d = \left\langle g_x,g_y,c_z\right\rangle
    \end{aligned}
\end{equation}

We consider \textit{ASME} as a paradigm that contains more autonomy than \textit{MSAE} because eversion presents fewer DoF than steering. 

\subsection{Mostly Autonomous}
\label{sec:mostly_autonomy}

In this paradigm, the operator has no control over the motion, and the manipulator's behavior is fully autonomous. However, as previously stated, the grasping operation and detection are still controlled by the operator. The execution of the task relies completely on the Goal Selection algorithm and the paradigm formulation is:

\begin{equation}\label{eq:fullAutonomy}
    \begin{aligned}
        d = \left\langle g_x,g_y,g_z\right\rangle
    \end{aligned}
\end{equation}


\section{Demonstrations and Results}
\label{sec:experiment}

In this section, we demonstrate the human operator teleoperation control of the soft-growing manipulation system and observe the performance and the human behavior under the different shared-control teleoperation paradigms introduced in Sec.~\ref{sec:shared_autonomy_paradigms}, while completing a pick-and-place manipulation task, described in Sec.~\ref{sec:task}. 
We let all the paradigms be tested by an expert and by a naive user for both real and simulated scenarios, evaluated the metrics given in Sec.~\ref{sec:evaluation_metrics}, and report the results in Sec.~\ref{sec:results} along with their discussion. 
%
Besides being used for evaluation, the simulated scenario was also used to perform a systematic parameters tuning procedure (involving a large number of trials) for the assistive algorithms (AAN and Fixed Assistance). Results of this study are given in Sec.~\ref{sec:params}.

\subsection{Task Description}
\label{sec:task}
During the execution of the task, the manipulator reaches for two items from above as shown in Fig.~\ref{fig:scheme}, and the operational workspace is a planar surface placed $700$~mm below the manipulator's base. The same task is performed for each shared-control teleoperation paradigm.

\begin{figure}[t]
  \centering
	{\includegraphics[width=0.6\linewidth]{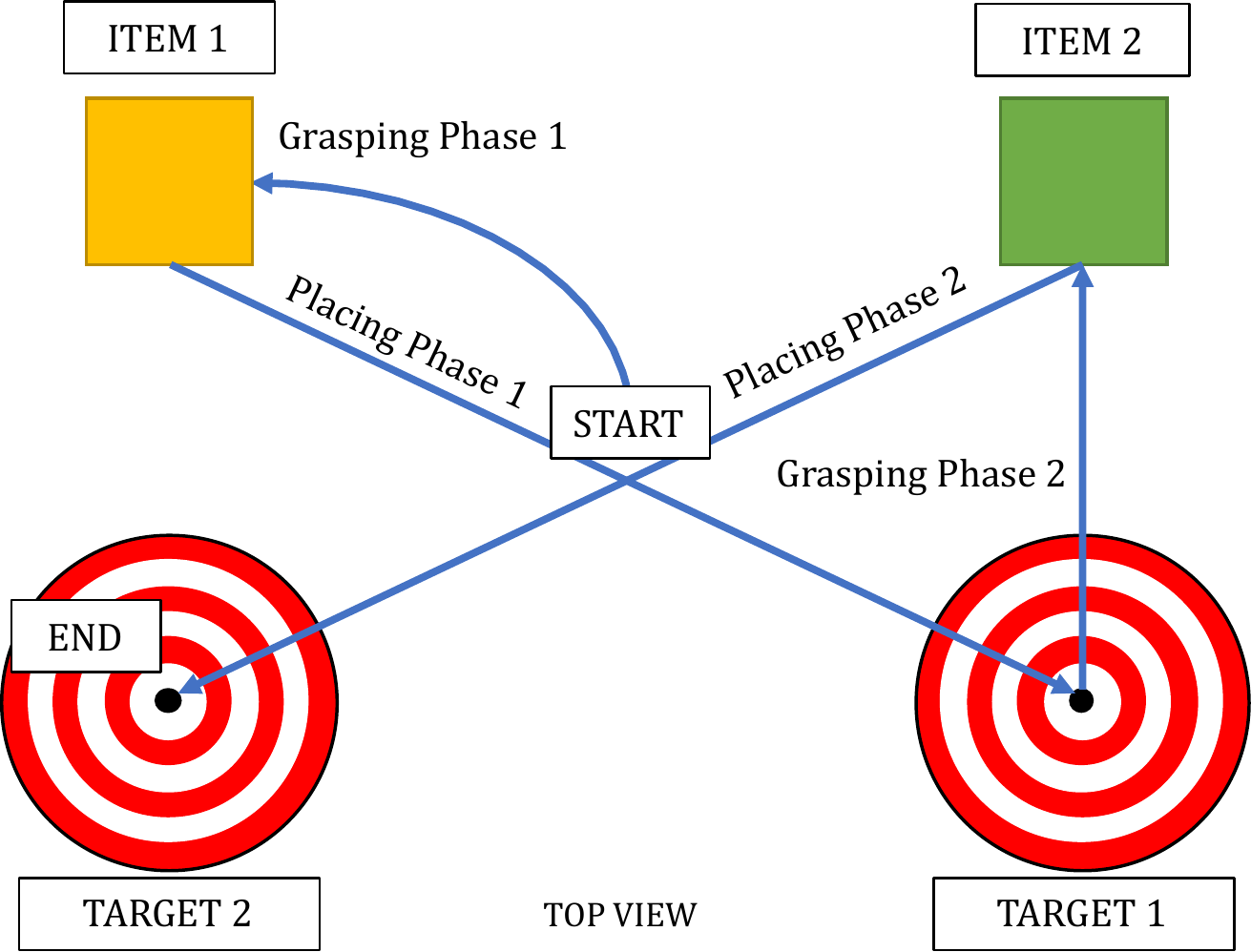}}\vspace{0.3cm}
    \caption{Progression of the task as structured in the workspace of the robot. Each trial starts with the manipulator at the center of the operational workspace. The grasping and placing phases are executed for item 1 and then item 2. The task ends when item 2 is placed on target 2. Arrows are for illustration purposes only and do not represent real paths or trajectories. The distances between items and targets are approximately 30~cm, and are not shown to scale in this figure. Each item is a foam cube with a side length of approximately 3~cm.}
    \label{fig:task_schematic}
\end{figure}

We define the following phases for each item, as shown in the state machine diagram in Fig.~\ref{fig:state_machine_GS}:
\begin{itemize}
    \item \textit{grasping phase}, in which the manipulator's objective is to reach and grasp an item (item is the goal); and 
    \item \textit{placing phase}, in which the manipulator's objective is to reach a target and release the current item on it (target is the goal).
\end{itemize}
For the pick-and-place of two items, a trial is thus composed of two repeated grasping and placing phases, as depicted in Fig.~\ref{fig:task_schematic}. The image shows the exact progression of a trial, which will be consistent in each scenario tested during the demonstrations. This means that the order of item selection is fixed, such that both the human and the robot have the same reference for execution. We observed that the layout of the goals does not affect the performance of the robot, and we defined the layout such that the robot would have to move in a large area of the operational workspace.

\subsection{Evaluation Metrics}
\label{sec:evaluation_metrics}

The following metrics were used to evaluate task performance: 

\begin{itemize}
    \item \textit{Item placement error}: the mean distance between the manipulator's end-effector and the target at the time of item release, projected on the $x$-$y$ plane of the operational workspace (mm);
    \item \textit{Execution time}: the time required to complete a trial (s); and
    \item \textit{Average amount of assistance}: the average force rendered on the haptic device to assist the operator, only for the \textit{Assist-as-Needed} and \textit{Fixed Assistance} paradigms (N/iterations)
    
\end{itemize}


\subsection{Parameter Tuning for Haptic Rendering}
\label{sec:params}
\textit{Assist-as-Needed} and \textit{Fixed Assistance} are the only paradigms whose parameters are directly affecting human behavior through haptic guidance. However, the assistive force rendered on the haptic device depends on several values as seen in Sec.~\ref{sec:aan} and~\ref{sec:fixed_assistance}, respectively. 
Thus, the effectiveness of these paradigms (both quantitative and qualitative)
depends on the choice of parameters. 
For \textit{Fixed Assistance} the choice can be straightforward, with only a few parameters to be set.

\begin{table}[b]
\small
\centering
\caption{Parameters values used for the systematic parameter tuning procedure performed for the Assist-as-Needed paradigm}
\label{tab:factors}
\begin{center}
 \renewcommand\arraystretch{1.2}
{
 \begin{tabular}{c | c c c c c c c c c}
\hline	
    Parameter & $f_{MAX}$ & $k_{MAX}$ & $\Delta$ & $\xi_S$ & $\xi_C$ & $\xi_A$ \\ \hline
    \texttt{low} & 3~N & 50~N/m & 2~s & 1~s & 1~s & 1~s \\
    \texttt{high} & 7~N & 100~N/m & 5~s & 3~s & 3~s & 3~s \\
\hline
\end{tabular} }
\end{center}
\end{table}

On the other hand, \textit{Assist-as-Needed} features multiple parameters, whose influence on the task performance may not be determined a priori, requiring a systematic parameter tuning method before demonstrations. The considered parameters were the maximum amount of force rendered ($f_{MAX}$), the maximum amount of stiffness used to generate the force ($k_{MAX}$), the increment in stiffness for each loop ($\Delta$), and the three reaction times to handle the cases of the operator moving close to the goal ($\xi_C$), away from the goal ($\xi_A$), or neither ($\xi_S$). 
For each parameter, we defined two levels: \texttt{low} and \texttt{high}. Table~\ref{tab:factors} shows the numeric values for each parameter.
We captured the aspects of the task performance with the following four performance metrics:

\begin{itemize}
    \item $\mathcal{T}$ completion time [s]: evaluated as the time from the start to the end of the trial - coincident with the second block being placed in the corresponding target zone;
    
    \item $\mathcal{L}$ trajectory length [m]: evaluated as the integral of the robot-tip velocity norm over time as in \ref{eq:anova_1};
    
    \begin{equation}
        \label{eq:anova_1}
        \mathcal{L} = \int_0^{\mathcal{T}} \dot{ee}(t) \, \textnormal{d}t
    \end{equation}
    
    \item $\mathcal{H}$ amount of assistance [N]: evaluated as the integral of the haptic guidance force norm over time divided by the completion time as in \ref{eq:anova_2}; and
    
    \begin{equation}
        \label{eq:anova_2}
	   \mathcal{H} = \frac{1}{\mathcal{T}}\int_0^{\mathcal{T}} \norm{\vec{f}}(t) \, \textnormal{d}t 
    \end{equation}
    
    \item $\mathcal{P}$ precision [m]: evaluated as the mean of the differences between the object and the target positions (projected into the horizontal plane) in the final configuration as in \ref{eq:anova_3};

    \begin{equation}
        \label{eq:anova_3}
        \mathcal{P} = \frac{1}{2}\sum^2_{i=1}\norm{t^{x,y} - p_i^{x,y}}
    \end{equation}
    
\end{itemize}

We evaluated these metrics along a set of trials performed accounting for all the possible parameters' combinations. $2^6 = 64$ trials were needed to test all the possible combinations. Each trial was repeated 3 times for a total of 192 trials: due to this high number, we decided to perform the assessment in the simulated environment presented in Sec.~\ref{sec:simulated}. 

\begin{table}[b]
	\small
	\centering
	\caption{Final choice of the parameters for haptic rendering in assistive paradigms}
	\label{tab:hapticRendering}
	\begin{center}
		\renewcommand\arraystretch{1.2}
		{
			\begin{tabular}{p{2.6cm}| c c }
				\hline		\multicolumn{2}{c}{\textit{Global}}		\\ \hline				
				\textit{refresh} & 100~Hz \\ 
				$f_{MAX}$ & 7~N  \\ 
				\hline		\multicolumn{2}{c}{\textit{Fixed Assistance}}		\\ \hline	
				$k$ & 10~N/m  \\ 
				$b$ & 0.1~Ns/m  \\
				${f}$ \textit{filter} & 50 samples  \\
				\hline		\multicolumn{2}{c}{\textit{Assist-as-Needed}}		\\ \hline	
				$\Delta$ & 20~Ns/m  \\ 
				$th_D$ & 30~mm  \\ 
				$th_M$ & 0.01~m/s  \\
				$\xi_S$ & 1~s  \\
				$\xi_C$ & 3~s  \\
				$\xi_A$ & 1~s   \\
				$k_{MAX}$ & 50~N/m \\
				\hline
		\end{tabular} }
	\end{center}
\end{table}

We ran an n-way ANOVA analysis on the results to identify significant parameters ($p < 0.05$ was considered statistically significant). We report the following findings:
\begin{itemize}
    \item  setting $f_{MAX}$ to \texttt{high} reduces the completion time of 18.184~s ($F(1,2) = 146.078$, $p < 0.01$), reduces the trajectory length of 0.197 m ($F(1,2) = 69.470$, $p = .014$), and increases the amount of assistance of 2.483 N ($F(1,2) = 22.415$, $p = .042$);
    \item setting both $\xi_{A}$ and $\xi_{S}$ to \texttt{low} reduces the completion time of 21.58~s ($F = 66.595$, $p = 0.015$) and the trajectory length of 0.251 m ($F(1,2) = 90.909$, $p = 0.011$) and of 0.395 m ($F(1,2) = 279.946$, $p < 0.01$), respectively;
    \item when $\xi_{S}$ is set to \texttt{high}, having $k_{MAX}$ to \texttt{low} increases the amount of assistance provided to the operator of 9.237 N ($F = 34.317$, $p = 0.028$) - stronger and more impulsive forces will be rendered to the operator; and
    \item the two-way interaction between $\xi_{C}$ and $\xi_{S}$ ($F(1,2) = 44.046$, $p = .022$) shows a higher increase in the amount of assistance when $\xi_{C}$ is set to \texttt{high} and $\xi_{S}$ is \texttt{low} - with these settings, a human operator is generally haptic-guided for a longer period of time as the assistive force raises rapidly in a steady state and decreases slowly when getting close to the target; and
    \item the precision is not significantly affected by the choice of the factors/levels.
\end{itemize}
We used these findings to tune the parameters accounting for our design objectives: aiming to reduce both completion time and trajectory length while having a high amount of assistance. The analysis suggests to set $f_{MAX}$ = \texttt{high}, $\xi_{A}$ = \texttt{low}, $\xi_{S}$ = \texttt{low}, $k_{MAX}$ = \texttt{low}, $\xi_{C}$ = \texttt{high}. The parameter $\Delta$ is free to be picked.
The final values we used in our setup are defined in Table~\ref{tab:hapticRendering}.

\subsection{Demonstrations}
To compare the performance of our shared-control teleoperation paradigms we performed multiple demonstration trials. Data shown below were collected from four distinct individuals (who introduced variability in the trials) teleoperating the system. Two individuals performed the tasks in the simulated environment and the other two performed the tasks in the real environment. In each environment, one participant was a naive user who had no previous experience in remotely controlling our soft-growing manipulator, and the other participant was an expert user -- one of the developers of the system. Each participant performed 5 trials of the two-item task for each of the 6 interaction paradigms, resulting in a total of 30 trials for each participant. 
The order of trials was randomized to prevent naive users from getting familiar with the control system as the demonstrations were carried out.
We did not observe trends in performance (i.e., learning curves) during 5 trials, although in pilot testing we found that learning curves were occasionally evident for naive users after 6 trials. Because we aimed to evaluate interaction paradigms for naive users before they had substantial experience, we did not go beyond 5 trials. Due to the substantial experience of the expert users while developing and tuning the simulated and real systems, their learning curves were assumed to have plateaued before the demonstration began. 

\subsection{Results and Discussion}
\label{sec:results}
The results from the demonstration trials are presented for each metric, showing average, median, interquartile range with outliers, and max/min across all trials for each paradigm. Charts and plots are shown in Fig.~\ref{fig:chart_errorAndTime},~\ref{fig:plot_errorVsTime} and~\ref{fig:chart_AAA}.

As in any remotely operated robotic system, the presence of time delay can affect system stability. We did not experience any instability during our experiments, but our paradigms could be combined with control strategies to address this (e.g., passivity-based control~\cite{nuno2011passivity}).

\subsubsection{Item Placement Error and Execution Time}

\begin{figure}[t]
\centering
    \subfigure[\protect\url{}\label{fig:chart_IPE}]%
	{\includegraphics[width=\textwidth]{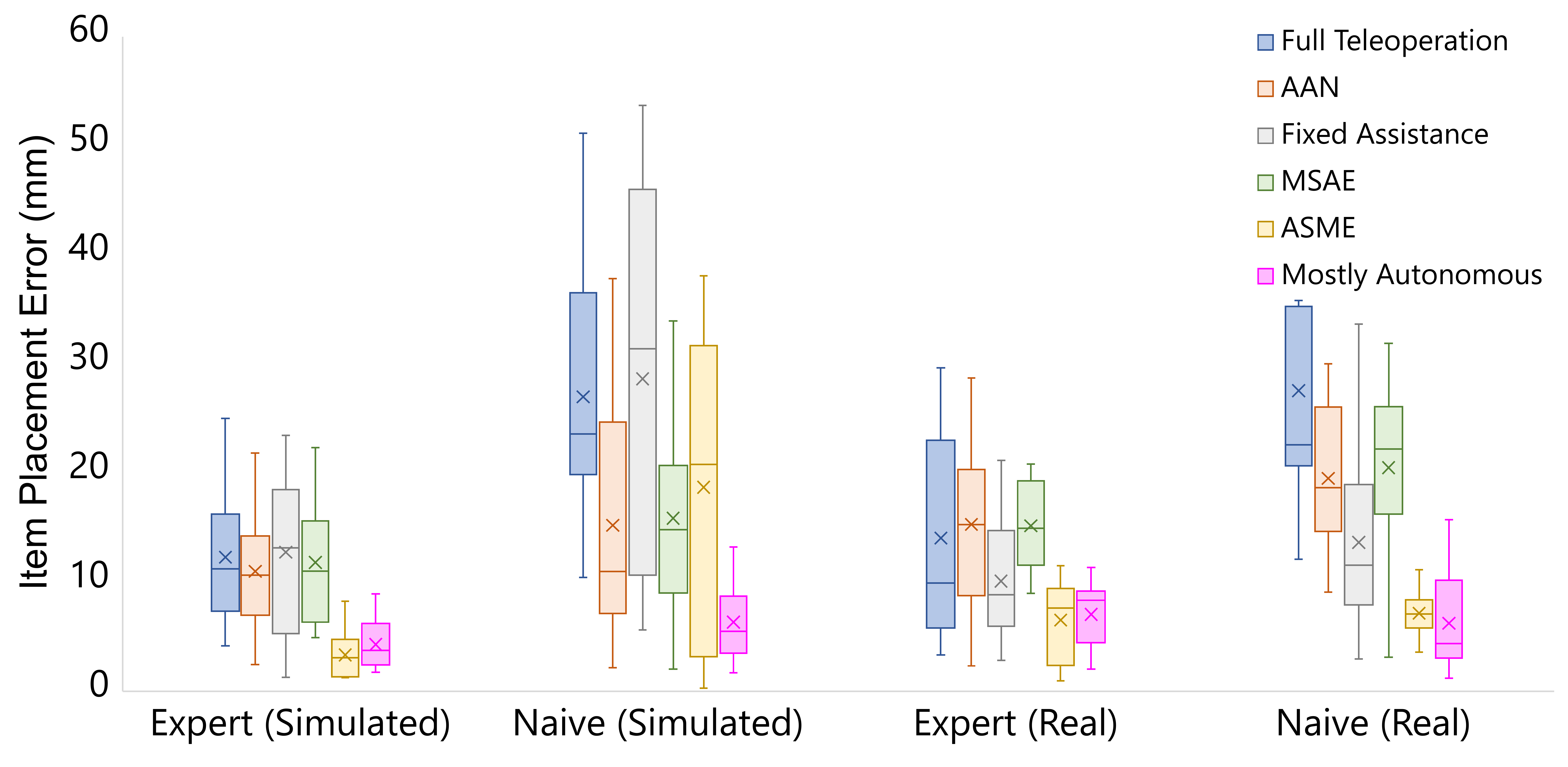}}
	\subfigure[\protect\url{}\label{fig:chart_ET}]%
	{\includegraphics[width=\textwidth]{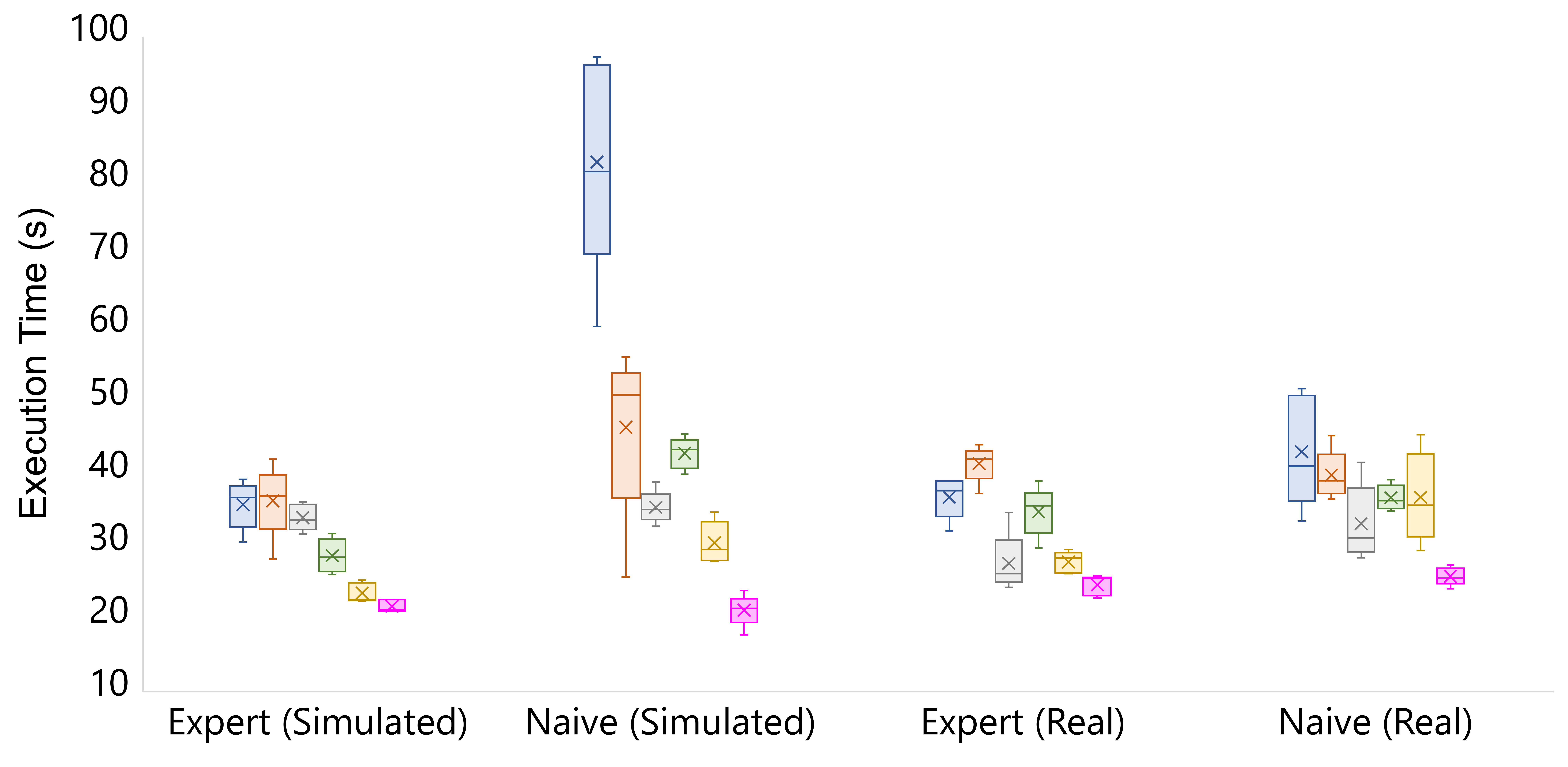}}\vspace{0.3cm}
	
	\caption{Results of different metrics for each participant. (a) Error in item placement. (b) Time of execution of the task.}\label{fig:chart_errorAndTime}
\end{figure}

\begin{figure}[t]
\centering
    \subfigure[\protect\url{}\label{fig:plot_errorVsTime_expert_sim}]%
	{\includegraphics[height=2.9cm]{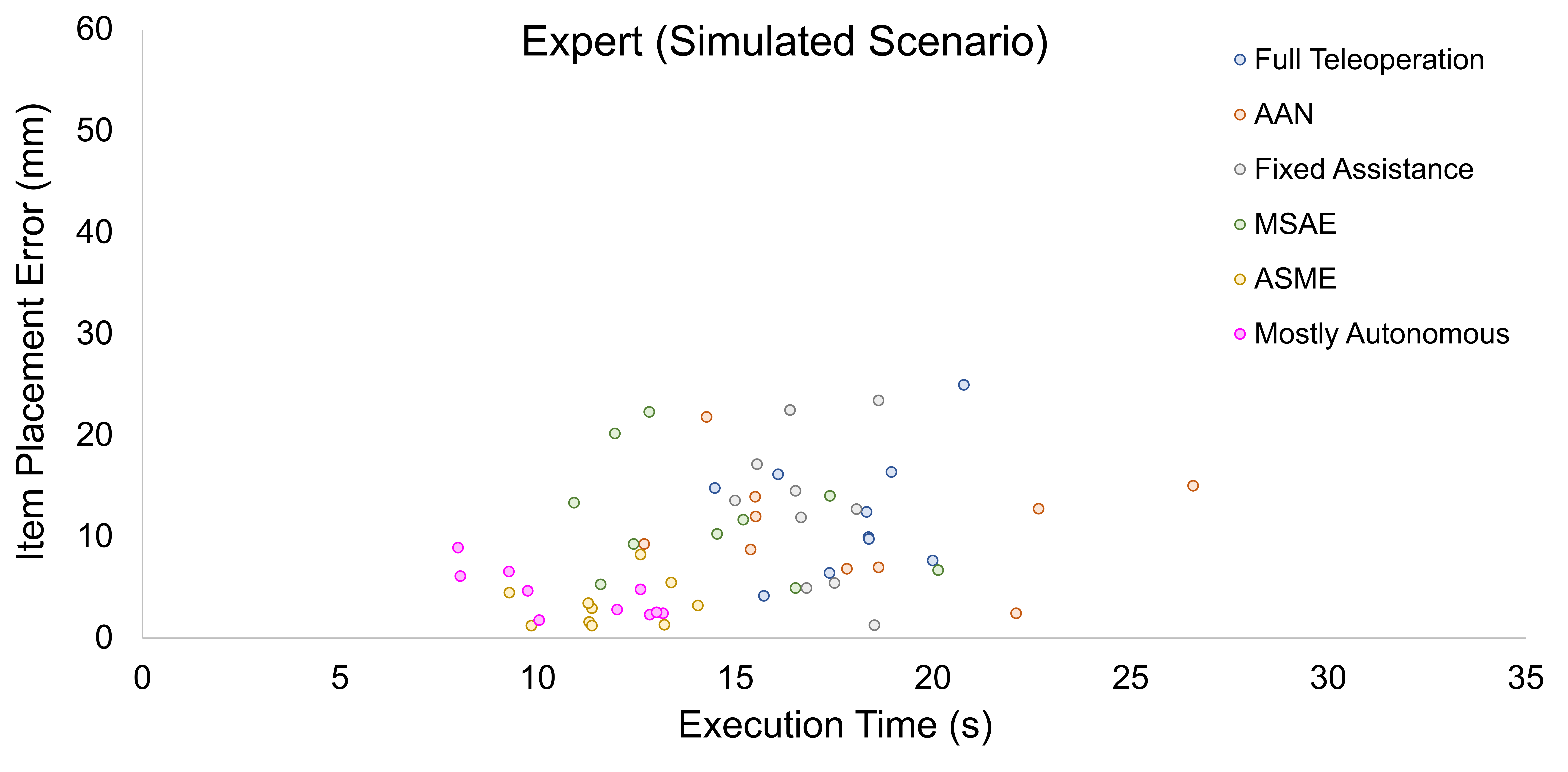}}
	\subfigure[\protect\url{}\label{fig:plot_errorVsTime_naive_sim}]%
	{\includegraphics[height=2.9cm]{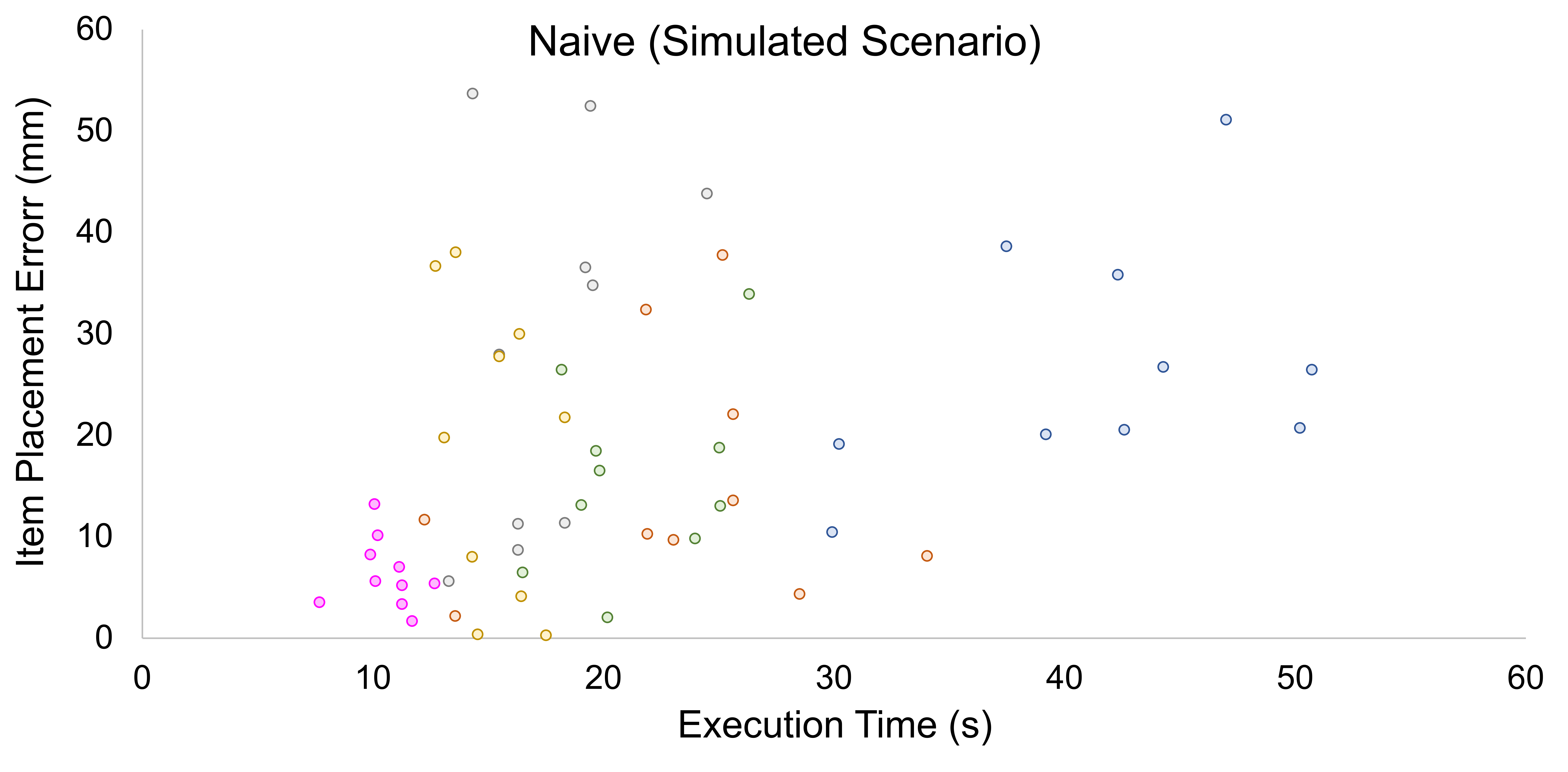}}
    \subfigure[\protect\url{}\label{fig:plot_errorVsTime_expert_real}]%
	{\includegraphics[height=2.9cm]{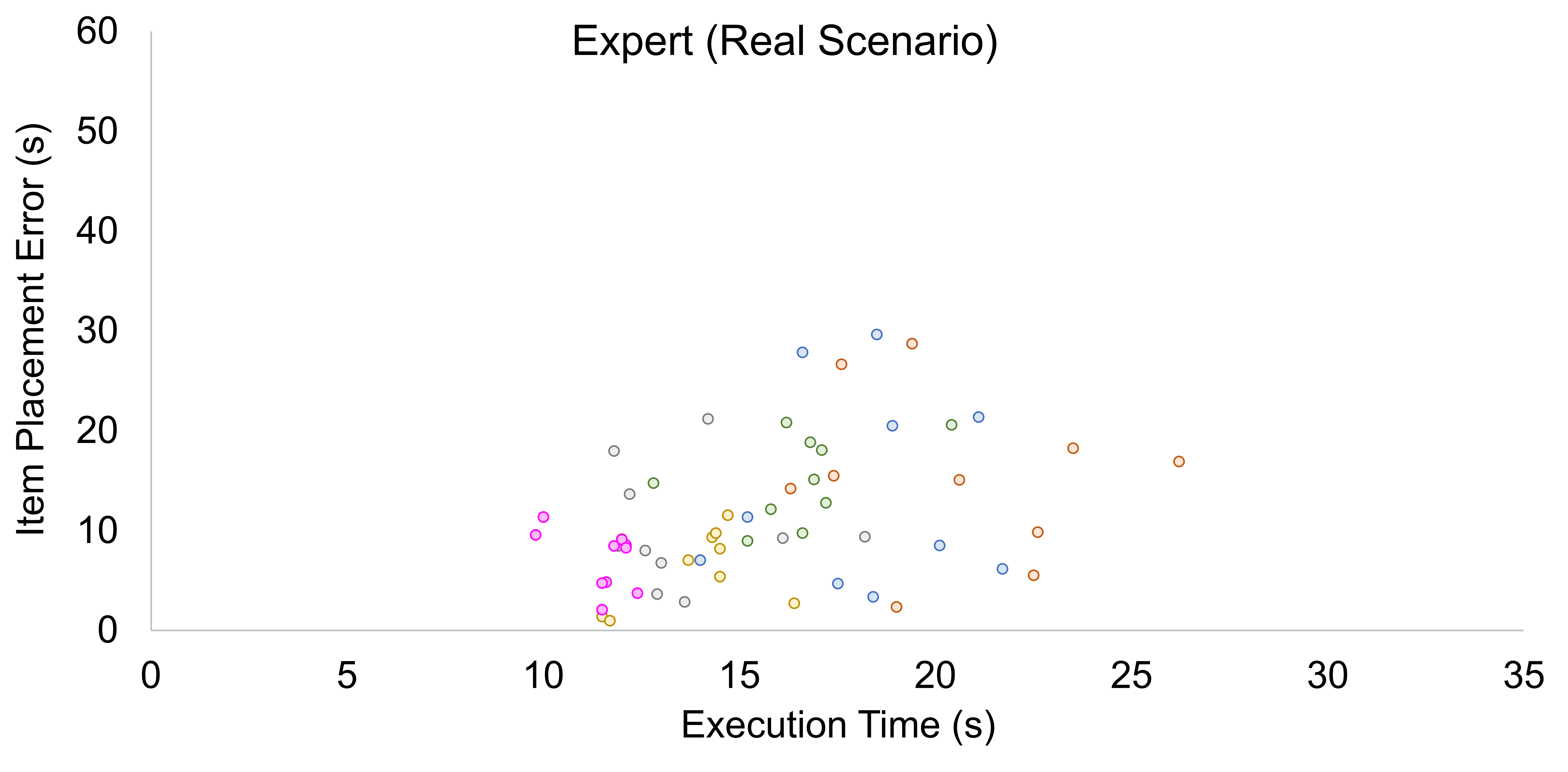}} 
	\subfigure[\protect\url{}\label{fig:plot_errorVsTime_naive_real}]%
	{\includegraphics[height=2.9cm]{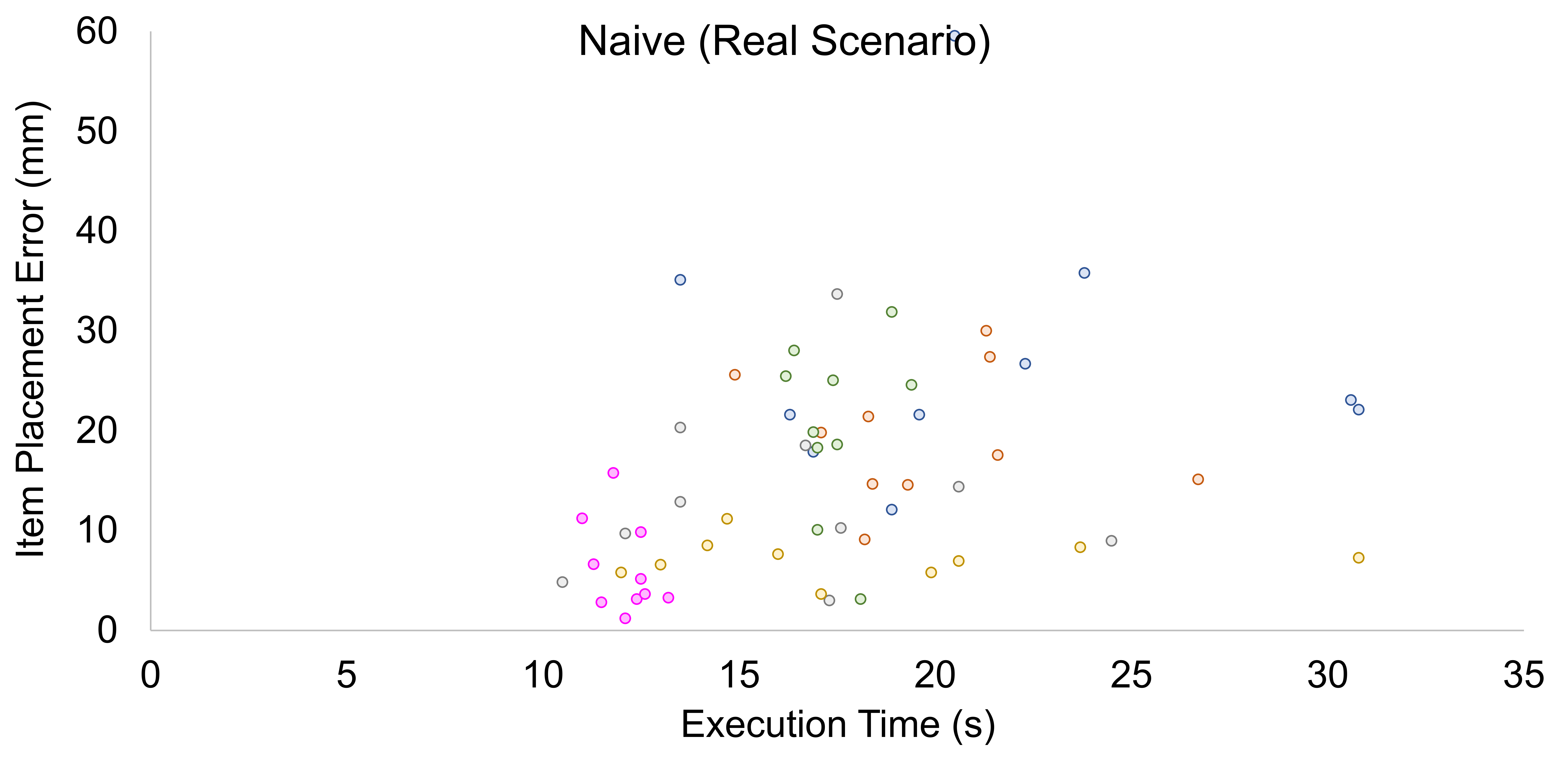}}\vspace{0.3cm}
	\caption{Error in item placement versus execution time for each paradigm, executed by both expert and naive users for both simulated and real scenarios. Here the time refers to the duration of the execution of grasping and placing phases of a single item. }\label{fig:plot_errorVsTime}
\end{figure}

 We compared item-placement error (Fig.~\ref{fig:chart_IPE}) and execution time (Fig.~\ref{fig:chart_ET}) against each other (Fig.~\ref{fig:plot_errorVsTime}) to emphasize performance distribution.
 The trend among all the paradigms is consistent: the best performances are achieved when the operator's control is limited.
 As expected, there is a visible difference between expert and naive users.

\textit{Full Teleoperation} and \textit{Assist-as-Needed} prioritize human control, where the operator's lack of perception is visible from the performance achieved in task execution.
This is noticeable in both the simulated and real scenario: in the first, 3D environments are difficult to navigate without stereoscopy or an extensive implementation of lightning and shadows; in the second, the distance between operator and device can proportionally affect perception~\cite{chen2007human}.

It is not surprising that \textit{Mostly Autonomous} achieved the shortest and most accurate trials, as the system always knows where the goal is. This trend is visible in the other semi-autonomous paradigms to different degrees, allowing for a shorter time than in \textit{Full Teleoperation} -- thus reducing the impairment caused by limited human perception. 
The values of item placement error obtained in \textit{Mostly Autonomous} are non-zero because the human operator still has full control over when and where to release items: higher values were obtained when the item was released while the manipulator was still approaching the target, both in simulated and real cases. Furthermore, the item-placing precision is dependent on several factors (e.g., release time, amount of pressure used to inflate the gripper, how the item is oriented when attracted to the gripper) such that, even when the system is automated, it is not entirely repeatable. Thus, the human operator plays an important role in manually counterbalancing undesired behaviors, emphasizing the importance of having the human in the loop.

Interestingly, \textit{MSAE} and \textit{ASME} do not show the same performance, with the latter being better in almost all cases. This finding might suggest that controlling steering is harder than controlling eversion: steering presents more challenges in actuation and control, as it is composed of more DoFs than eversion.
We hypothesize that the entity (human or robot) controlling the steering has  more control over the system, and as such \textit{ASME} is a more automated paradigm than \textit{MSAE}.

\begin{figure}[h]
  \centering
	{\includegraphics[width=0.8\linewidth]{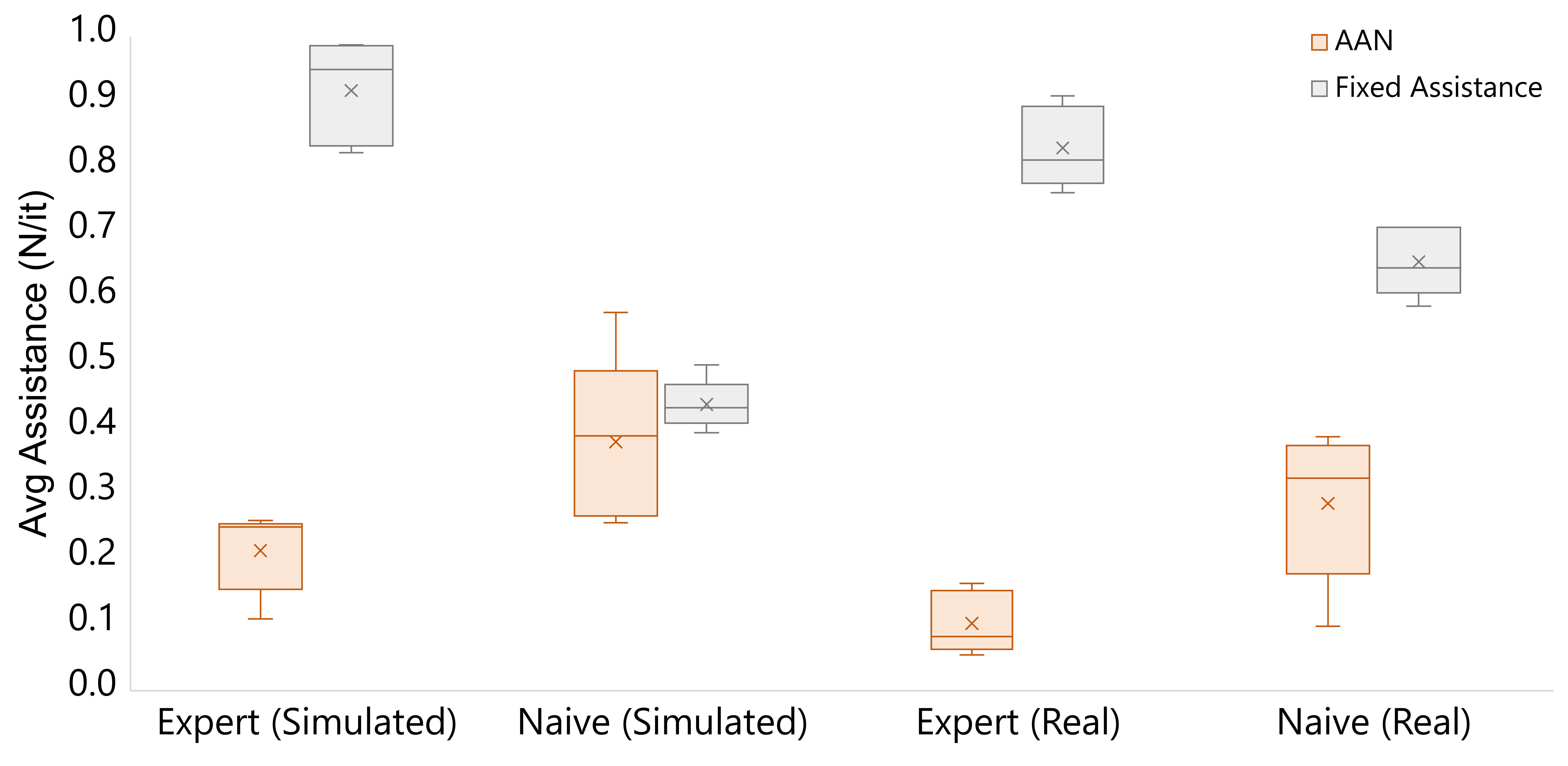}}\vspace{0.3cm}
    \caption{Average amount of assistance rendered on the haptic interface in the assistive paradigms, executed by both expert and naive user in the simulated and the real scenario.}
    \label{fig:chart_AAA}
\end{figure}

\textit{Fixed Assistance} is similar to \textit{MSAE}, with results in the same range for expert users, and slightly worse for naive ones. 
This suggests that naive users might be deceived by constant haptic feedback (Fig.~\ref{fig:chart_AAA}), which can be hard to accommodate when the operator is not used to it. Indeed, naive users performed slightly better in \textit{Assist-as-Needed} than in \textit{Fixed Assistance}, especially in the simulated scenario. 

Finally, little difference was found between \textit{Full Teleoperation} and \textit{Assist-as-Needed} for experts. As shown in Fig.~\ref{fig:chart_AAA}, the level of haptic assistance for \textit{Assist-as-Needed} was barely perceivable because expert users did not need any assistance -- they were highly experienced in teleoperating the soft-growing manipulator. This finding is also observed in Fig.~\ref{fig:plot_errorVsTime}: data for \textit{Full Teleoperation} and \textit{Assist-as-Needed} performance are similar for expert users (Fig.~\ref{fig:plot_errorVsTime_expert_sim} and \ref{fig:plot_errorVsTime_expert_real}), but different for naive users, with \textit{Assist-as-Needed} resulting in slightly better performance (Fig.~\ref{fig:plot_errorVsTime_naive_sim} and \ref{fig:plot_errorVsTime_naive_real}). The difference between expert and naive users in the \textit{Assist-as-Needed} paradigm shows that the method works as intended: the assistance is given and perceived only when the task was not performed correctly; expert users did not need it, and thus their results were not different from the \textit{Full Teleoperation} case, whereas naive users occasionally required more guidance.

\subsubsection{Participant's Opinions}
In addition to the numerical results, we also considered how the task execution was perceived by the operators in the different interaction paradigms. Participants were asked questions to evaluate their role during the tasks. All the participants agreed that the task was difficult to accomplish under \textit{Full Teleoperation}, especially in phases requiring accuracy such as when approaching targets. \textit{Assist-as-Needed} was perceived differently, with naive users being correctly relieved by occasional assistance and experts barely noticing any. The opposite reaction was reported for \textit{Fixed Assistance}, where experts found it a useful tool for guidance while naive users found it too constraining and hard to control. \textit{MSAE} and \textit{ASME} were both rated better than \textit{Full Teleoperation} and described as the most collaborative paradigms in terms of interaction with the robot. \textit{ASME} was rated easier to control but somewhat less interactive. Finally, \textit{Mostly Autonomous} was rated as the easiest by all participants, enjoyable to observe but not very compelling in terms of interaction. 

\subsubsection{Paradigms for Soft-Growing Robots}

The differences in performance between the \textit{MSAE} (with the user controlling steering) and \textit{ASME} (with the user controlling growth) paradigms suggest that the soft-growing robot's unique decoupling of DoFs can enable unique combinations of manual and autonomous control, expanding the options for shared control. In our results, \textit{ASME} performed generally better than \textit{MSAE}, allowing users to remain in the control loop of the robot with the possibility for increased situational awareness, while still allowing the autonomous algorithms to handle more challenging components of the task.
These results suggest that sharing the control with semi-autonomous soft-growing robots increases the performance of manipulation tasks, not only in terms of the analyzed metrics but also in terms of maneuverability perceived by the operator. 


\section{Conclusion}
\label{sec:conclusion}

In this paper, we introduced, validated, and compared the performance of six shared-control teleoperation paradigms with different autonomy and assistance levels on a soft-growing robot manipulator. 
The proposed paradigms were designed so as to explore a wide spectrum of potential operation modes going from full teleoperation to quasi-full autonomy, gradually reducing the role of the human operator in favor of robot autonomy. 
Results show that, by gradually adding autonomy to basic teleoperation, the performance improves to the detriment of human involvement. 
We noticed that having constant haptic guidance is a better paradigm for expert users, whereas having need-tailored guidance is preferred when users are new to the task. 
Having the human in control of partial DoFs might be the best solution in terms of perceived involvement and performance, whereas limiting the human control is useful when the task requires extreme precision and the system is capable of acquiring enough information from the environment to perform a correct execution. Wherever achieving this precision is not possible, the role of the human is to fine-tune position when close to targets, leaving the navigational part to the robot.


In the future, additional interaction paradigms can be developed and tested based on novel features of not only soft-growing robots but other soft robot manipulators with novel kinematics. These can be blended together to obtain a mixture of shared-control paradigms, which would lead to a continuous rather than a discrete set of control modalities that could be desirable in some applications. In this way, the user could select from a continuous range the percentage of manual versus autonomous control along each DoF of the system. For instance, considering a linear blending strategy, a real parameter can be used to weigh the control inputs coming from an autonomous controller and the human or from two different controllers along the same DoF. Blending and transitioning among control strategies could result in unexpected emergent behaviors of the system that affect usability, and should be studied carefully in future work.
Additionally, we are interested in developing new paradigms exploiting the softness of our manipulator, allowing for collaborative tasks in which robots are not harmful to humans or the environment in case of undesired collisions.

Finally, due to the COVID-19 pandemic restricting human subjects' work as well as the very large number of conditions tested in the targeted user studies presented here, we had selected to demonstrate the system with only one experienced and one naive user in each scenario rather than running a complete user study. In future work, we will perform a larger user study to understand the effects of our interaction modes in scenarios that can be more complex than the one proposed in this pilot work (e.g., exploration of cluttered environments) and with further limitations (e.g., limited perception of the robot to emphasize the role of the human operator).

\backmatter

\bmhead{Supplementary information}

The accompanying supplementary video shows simulated and real demonstrations carried out in this work, whereas the accompanying supplementary document describes the details of the inverse kinematics and control of the soft-growing manipulator.

\bibliography{references}


\section*{Statements and Declarations}

\bmhead{Funding} 
This work was supported in part by Toyota Research Institute (TRI) and National Science Foundation grant 2024247. TRI  provided funds to assist the authors with their research but this article solely reflects the opinions and conclusions of its authors and not TRI or any other Toyota entity.

\bmhead{Competing interests}
The authors have no relevant financial or non-financial interests to disclose.

\bmhead{Authors' contributions}

Fabio Stroppa designed and developed the interaction paradigms, designed and developed the overall system (motion tracking, computer vision, network communication, interfaces), designed and developed the experimental setup, designed and built the circuit board for the soft robot, assembled the soft robot, designed and implemented the inverse kinematics for the soft robot, supervised the real-scenario demonstrations, analyzed the data, and wrote the manuscript.
Mario Selvaggio designed and implemented the virtual model of the soft robot for simulation, ran the statistical analysis for the parameter tuning for haptic rendering, supervised the simulated-scenario demonstrations, and partially wrote the manuscript.
Nathaniel Agharese and Laura H. Blumenschein contributed to implementing the closed-loop control of the system and reviewed the manuscript.
Ming Luo built the prototype of the soft robot and reviewed the manuscript.
Elliot W. Hawkes reviewed the manuscript.
Allison M. Okamura supervised the project and reviewed the manuscript.

\bmhead{Ethics approval}
This is an observational study. The Institutional Review Board of Stanford University has confirmed that no ethical approval is required.

\bmhead{Consent to participate}
Informed consent was obtained from all individual participants included in the study

\bmhead{Consent for publication}
No individual person’s data in any form is disclosed in the paper.

\bmhead{Code or data availability} 

The simulator scene and its scripts, as well as the C++ software implementing the interaction modules described in the text, are publicly released at \url{https://github.com/mrslvg/sgm-sim/}.

\end{document}


\title[Shared-control Teleoperation Paradigms on a Soft-Growing Robot Manipulator]{\bf Soft-Growing Robot Manipulator\\ Kinematics and Control}

\maketitle



This document briefly illustrates the strategies used to control the soft-growing robot, including details on the inverse kinematics and the closed-loop control.

\section{Inverse Kinematics}
\label{sec:inverseKinematics}

\begin{figure}[h]
\centering
    \subfigure[\protect\url{Cartesian}\label{fig:ik_mapping_c}]%
	{\includegraphics[height=3.3cm]{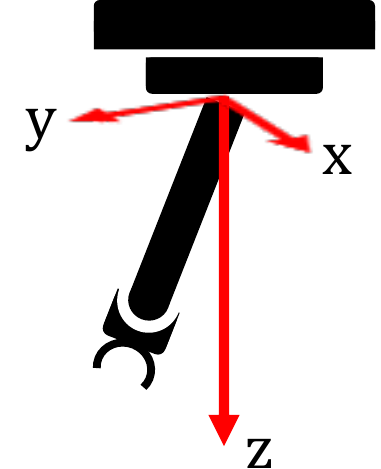}}\qquad
	\subfigure[\protect\url{Motor}\label{fig:ik_mapping_m}]%
	{\includegraphics[height=3.3cm]{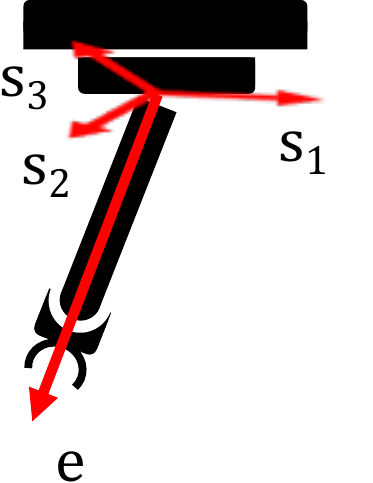}}\vspace{0.3cm}
	\caption{The inverse kinematics transform a point from the (a) Cartesian space into the (b) manipulator motor space, defined by three dimensions for steering in the plane and a fourth for eversion. Eversion is oriented with the body of the robot.}
	\label{fig:ik_mapping}
\end{figure}

The controller includes inverse kinematics that transform a point from Cartesian space into motor space, from three coordinates $\langle x , y, z \rangle$ to four $\langle s_1 , s_2, s_3, e \rangle$ as depicted in Fig.~\ref{fig:ik_mapping}. 
In motor space, the dimensions $s_1$, $s_2$, and $s_3$ define the direction of the three motors used for steering in the plane, whereas $e$ is used for eversion. 

The procedure is divided into mapping for steering and eversion, which are described in the following sections.
We solve the inverse kinematics with a geometric rather than differential approach because the inherent difference in steering and eversion dynamics required unique gains for the motors controlling the different DoFs.

\section{Steering Mapping}

\begin{figure}[h]
\centering
    \subfigure[\protect\url{}\label{fig:steeringMapping_cartesian}]%
	{\includegraphics[height=3cm]{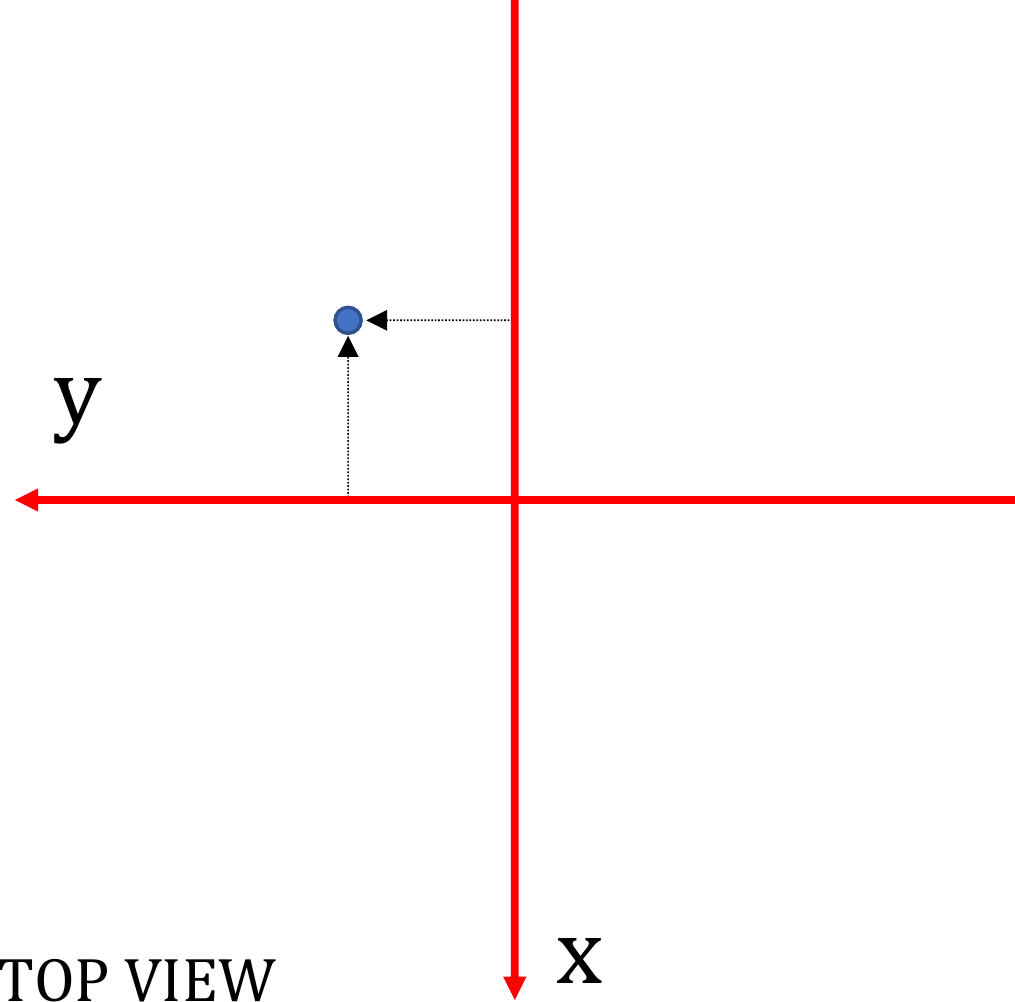}}
	\subfigure[\protect\url{}\label{fig:steeringMapping_motor}]%
	{\includegraphics[height=3cm]{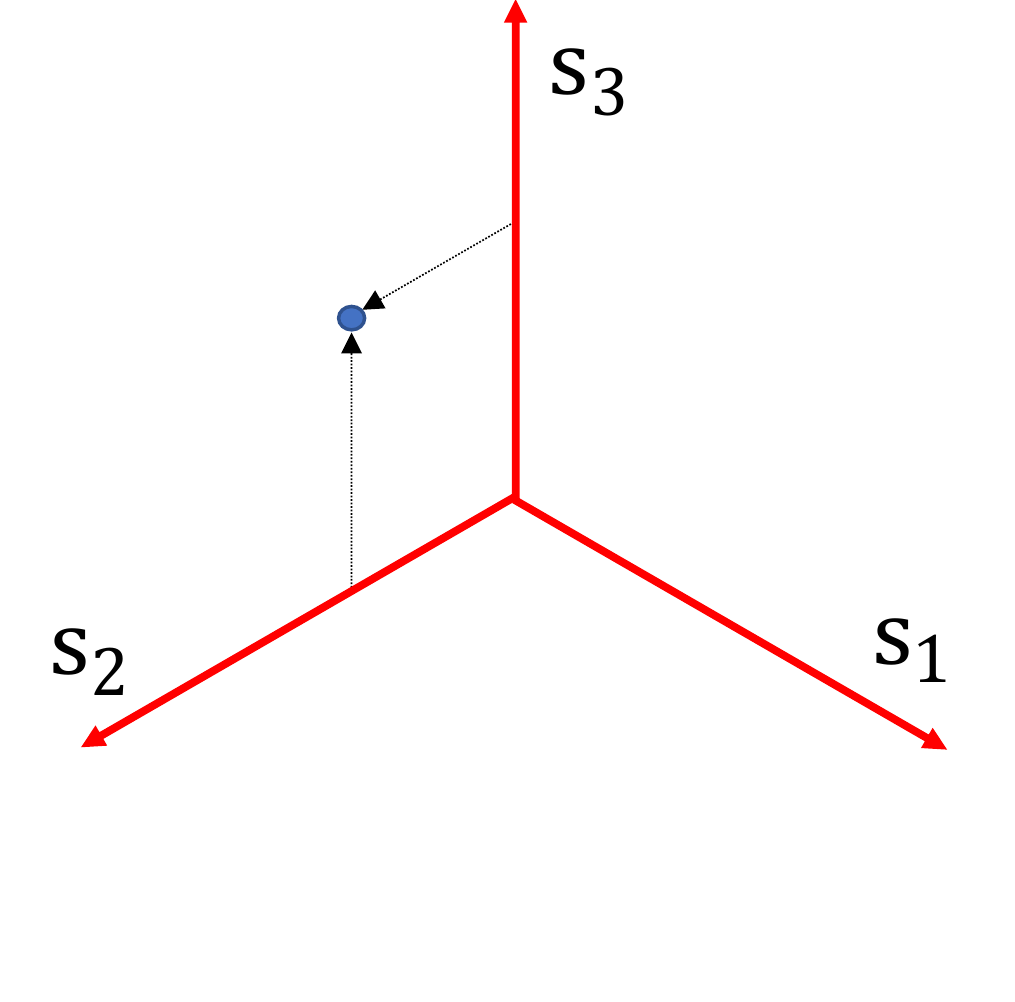}}
	\subfigure[\protect\url{}\label{fig:steeringMapping_mixed}]%
	{\includegraphics[height=3cm]{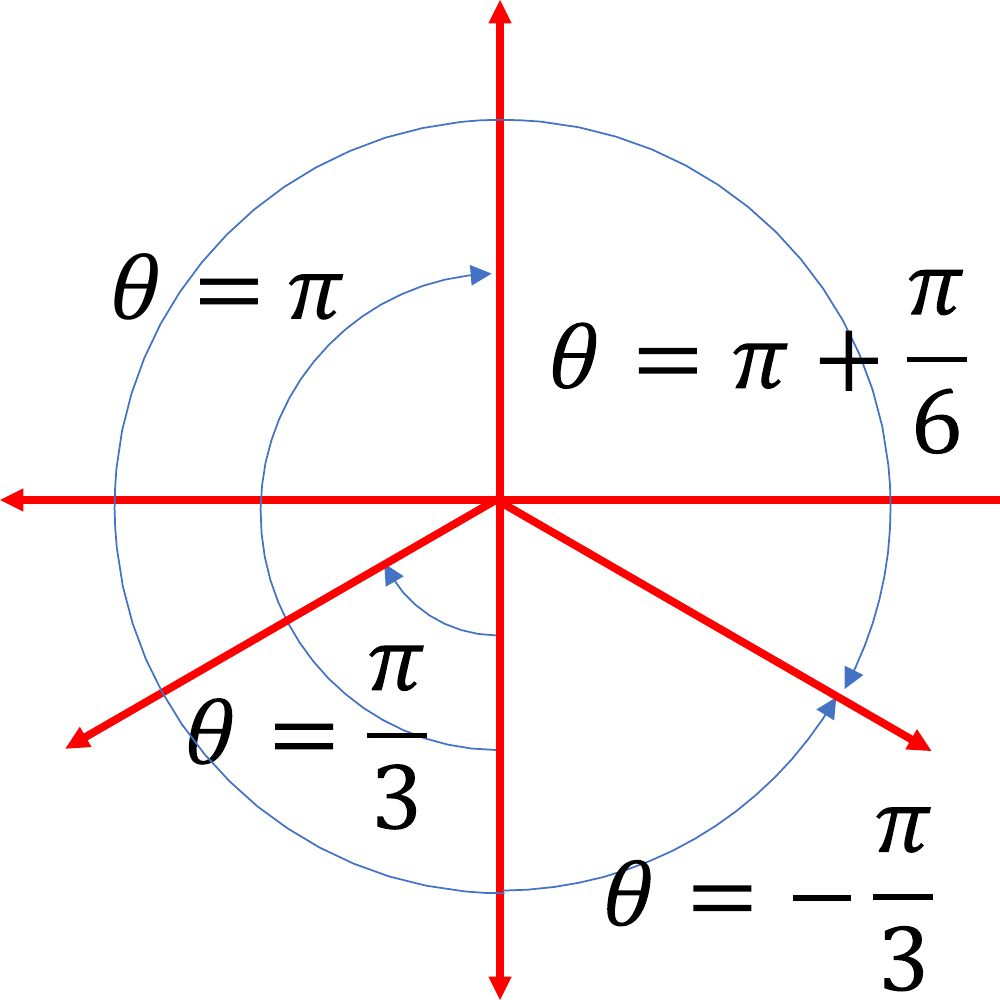}}\vspace{0.3cm}
	\caption{(a) A point in Cartesian space. (b) The same point expressed in motor space. (c) One representation of the angles $\theta$ between the $x$-axis of the Cartesian space and the axes of the Motor space. These angles are used to identify in which sector the point lies, then solve the inverse kinematics for those two dimensions defining that sector by setting the third dimension to zero.}
	\label{fig:steeringMapping}
\end{figure}

The mapping for steering transforms a point in a plane from a 2D representation in Cartesian space ($\langle x , y \rangle$ as shown in Fig.~\ref{fig:steeringMapping_cartesian}) to a 3D representation in Motor space ($\langle s_1 , s_2, s_3  \rangle$  as shown in Fig.~\ref{fig:steeringMapping_motor}). 
The same plane composed of four quadrants in the Cartesian space is divided into three sectors by the axes of the motor space, having an angle of $\frac{2}{3}\pi$ between each pair. 
These axes are aligned with the direction of the manipulator's motors: the positive direction of the axes indicates a rotation of the motor to retract the cable and pull the manipulator, whereas the negative direction indicates cable release. A coordinated movement along these axes allows the manipulator to steer in a desired direction.
The mapping is defined by the following equations, which are given by expressing the half lines of the motor space axes in Cartesian space.
\vspace{-0.2cm}
\begin{eqnarray}\label{eq:steeringSystem}
        x & = & s_1 \cdot \cos{\left(-\frac{\pi}{3}\right)} + s_2 \cdot \cos{\left(\frac{\pi}{3}\right)} - s_3  \nonumber \\
        y & = & s_1 \cdot \sin{\left(-\frac{\pi}{3}\right)} + s_2 \cdot \sin{\left(\frac{\pi}{3}\right)}
\end{eqnarray}

\noindent Having a system of two equations in three variables leads to infinite solutions; therefore, we fix one motor to be zero such that only two motors are actuated at a time. 
The motor set to zero is the one along the axis that does not define the sector in which the point currently lies: by defining $\theta$ as the angle between the positive $x$-axis and the ray to the point $\langle x, y \rangle$ (Fig.~\ref{fig:steeringMapping_mixed}), the solution to ~\eqref{eq:steeringSystem} is given by the following:

\begin{itemize}

\item When the point lies in the sector defined by $s_1$ and $s_2$, the coordinate $s_3$ is set to zero, and the solution is:
\begin{equation}\label{eq:solution_tridant_1}
    \mbox{if } \theta \in \left[ -\frac{\pi}{3}, \frac{\pi}{3} \right]
    \begin{cases} 
        s_1 = x \cdot \frac{1}{2\cos{(\frac{\pi}{3})}} - y \cdot \frac{1}{2\sin{(\frac{\pi}{3})}} \\ 
        s_2 = x \cdot \frac{1}{2\cos{(\frac{\pi}{3})}} + y \cdot \frac{1}{2\sin{(\frac{\pi}{3})}} \\
        s_3 = 0
    \end{cases}
\end{equation}

\item When the point lies in the sector defined by $s_2$ and $s_3$, the coordinate $s_1$ is set to zero, and the solution is:
\begin{equation}\label{eq:solution_tridant_2}
    \mbox{if } \theta \in \left[ -\frac{\pi}{3}, \pi \right]
    \begin{cases} 
        s_1 = 0 \\ 
        s_2 = y \cdot \frac{1}{2\sin{(\frac{\pi}{3})}} \\
        s_3 = y \cdot \frac{1}{2\tan{(\frac{\pi}{3})}} - x
    \end{cases}
\end{equation}

\item When the point lies in the convex sector defined by $s_1$ and $s_3$, the coordinate $s_2$ is set to zero, and the solution is:
\begin{equation}\label{eq:solution_tridant_3}
    \mbox{if } \theta \in \left[ \pi, \pi + \frac{\pi}{6} \right]
    \begin{cases} 
        s_1 = - y \cdot \frac{1}{2\sin{(\frac{\pi}{3})}} \\ 
        s_2 = 0 \\
        s_3 = - x - y \cdot \frac{1}{2\tan{(\frac{\pi}{3})}}
    \end{cases}
\end{equation}
\end{itemize}

\begin{figure}[h]
  \centering
	{\includegraphics[width=6cm]{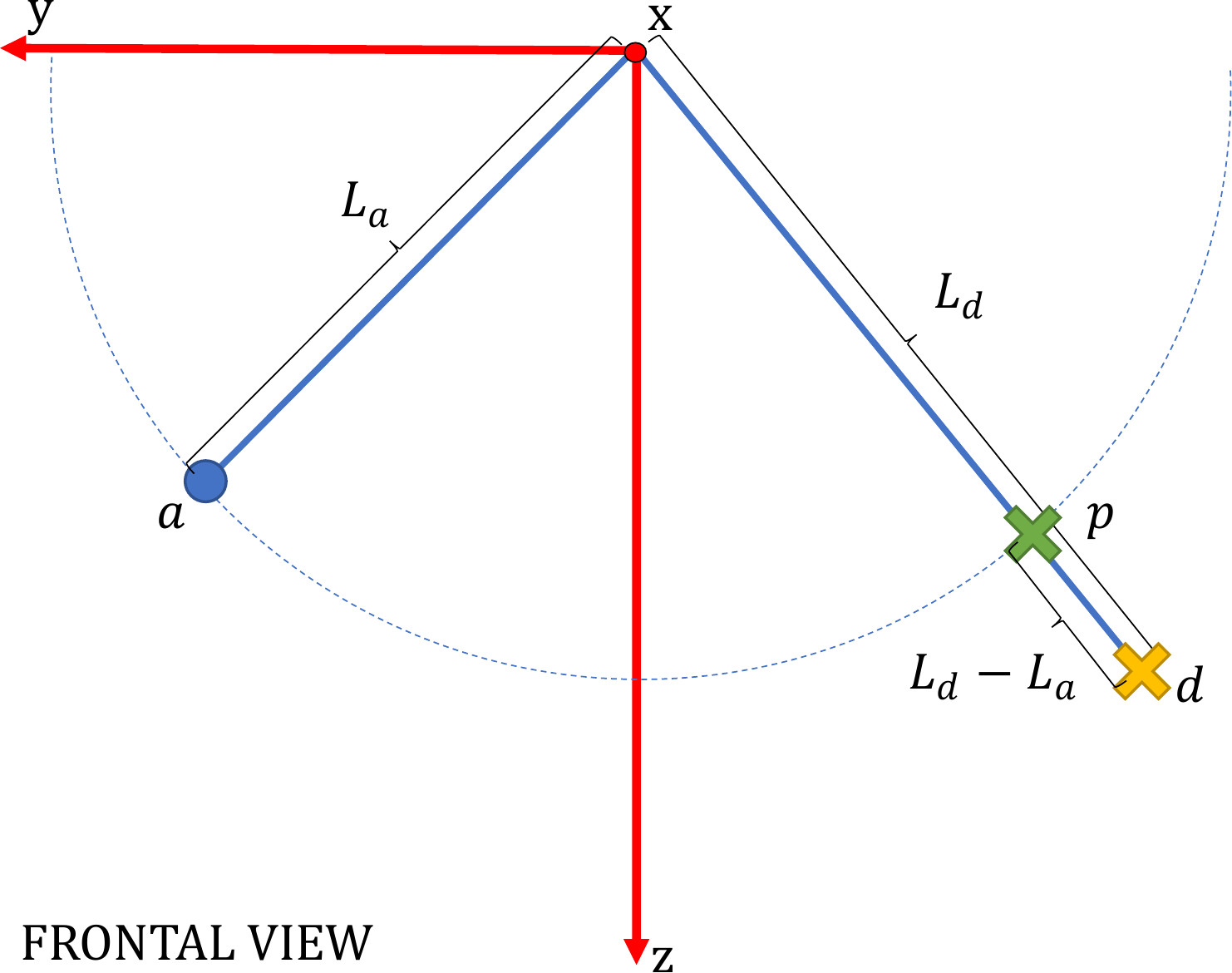}}\vspace{0.3cm}
    \caption{The inverse kinematics calculates the steering commands on a proxy $p$ rather than the real desired position $d$, to decouple steering from eversion. }
    \label{fig:proxy_kinematics}
\end{figure}

\section{Eversion Mapping}

Starting from a point in the Cartesian space, the eversion coordinate $e$ is simply given by the length of the robot calculated as in~\eqref{eq:eversionCoordinate}.
\begin{equation}\label{eq:eversionCoordinate}
    e  = \sqrt{x^2 + y^2 + z^2}
\end{equation}

\section{Error Evaluation} \label{Error evaluation}

We calculate the error between the the desired position, $d = \langle x , y , z \rangle \in \mathbb{R}^3$, and the actual position, $a = \langle x , y , z \rangle \in \mathbb{R}^3 $, of the robot end-effector for closed-loop control. The output error is $\psi = \langle s_1 , s_2 , s_3 , e \rangle \in \mathbb{R}^4$.
The inputs $d$ and $a$ are defined in Cartesian space, whereas the output $\psi$ is in motor space.
To solve the system in~\eqref{eq:steeringSystem}, we constrain one of the steering dimensions to be zero to ensure balance among cable actuation.
In the following steps, we reference the same point ($a$ or $t$) in the two different spaces: when this is accompanied by the superscript $m$, then we refer to motor space; if no superscript is indicated, then we refer to Cartesian space.

We first calculate the length of the robot, for both the actual ($L_a$) and the desired ($L_d$) configuration. 
%
%
Next, in order to decouple steering and eversion, the commands in the plane are evaluated on a proxy rather than directly on the desired position $d$. 
This proxy $p$ is the position the manipulator would reach without eversion, which is achieved by steering from $a$ to $d$ following the surface of a semi sphere, as shown in Fig.~\ref{fig:proxy_kinematics}. 
Once $p$ has been given as a goal for steering, the position $d$ is reached by everting the manipulator of the required length $L_d - L_a$.

The proxy $p$ is obtained by multiplying a unit vector $\hat{u_d}$ pointing in the direction of $d$ and the displacement in length required to keep the manipulator on the same semi sphere, which is the actual length of the robot $L_a$. 
    \begin{equation}\label{eq:proxySphere_1}
        \hat{u}_d = \frac{d}{\abs{d}} = \frac{d}{L_d} = \frac{d}{\sqrt{{d_x}^2+{d_y}^2+{d_z}^2}}
    \end{equation}
    \begin{equation}\label{eq:proxySphere2}
        p = \hat{u}_d \cdot \sqrt{{a_x}^2+{a_y}^2+{a_z}^2} = \hat{u}_d \cdot L_a = d \cdot \frac{L_a}{L_d}
    \end{equation}


The steering actuation is primarily achieved by pulling cables; releasing a cable will still actuate the robot, but the effect is reduced as no force is exerted against the friction of the cable routing except for gravity. Because cable release does not  respect the inverse kinematics, we perform all steering as cable pulling as follows. First, $p$ and $a$ are translated such that $a$ coincides with the origin of the reference frame, generating a point $t$. Next, the point $t$ is transformed into motor space $t^m$, which will only have positive commands (cable pulling) to ensure actuation.

Unfortunately, performing only cable pulling results in wrinkling of the fabric of the robot at the base where the steering occurs. Too much wrinkling leads to the robot being unable to steer. Thus, when the robot is not required to steer in the positive direction of a given motor, that motor rotates in the opposite direction avoiding wrinkling. In our kinematics, this translates into transferring part of the most positive value from one steering axis to the other two in the opposite direction; or in other words, finding a different route in the space for reaching the same desired point from a starting one. Finding a different route is always possible, as the system in~\eqref{eq:steeringSystem} has infinite solutions. 

The most positive value $t^m_{s_{max}}$ is defined by~\eqref{eq:redistribuition_math}, and is it always possible to retrieve it as we chose the solutions~\eqref{eq:solution_tridant_1},~\eqref{eq:solution_tridant_2}, and~\eqref{eq:solution_tridant_3} to be always positive. 
\begin{align}\label{eq:redistribuition_math}
    \forall j \in [1,3] \enskip : \enskip 0 \leq t^m_{s_j} \leq t^m_{s_{max}} = \max t^m_{s_j} 
\end{align}

\begin{figure}[h]
\centering
    \subfigure[\protect\url{}\label{fig:Cable_Geometry2_A}]%
	{\includegraphics[height=4cm]{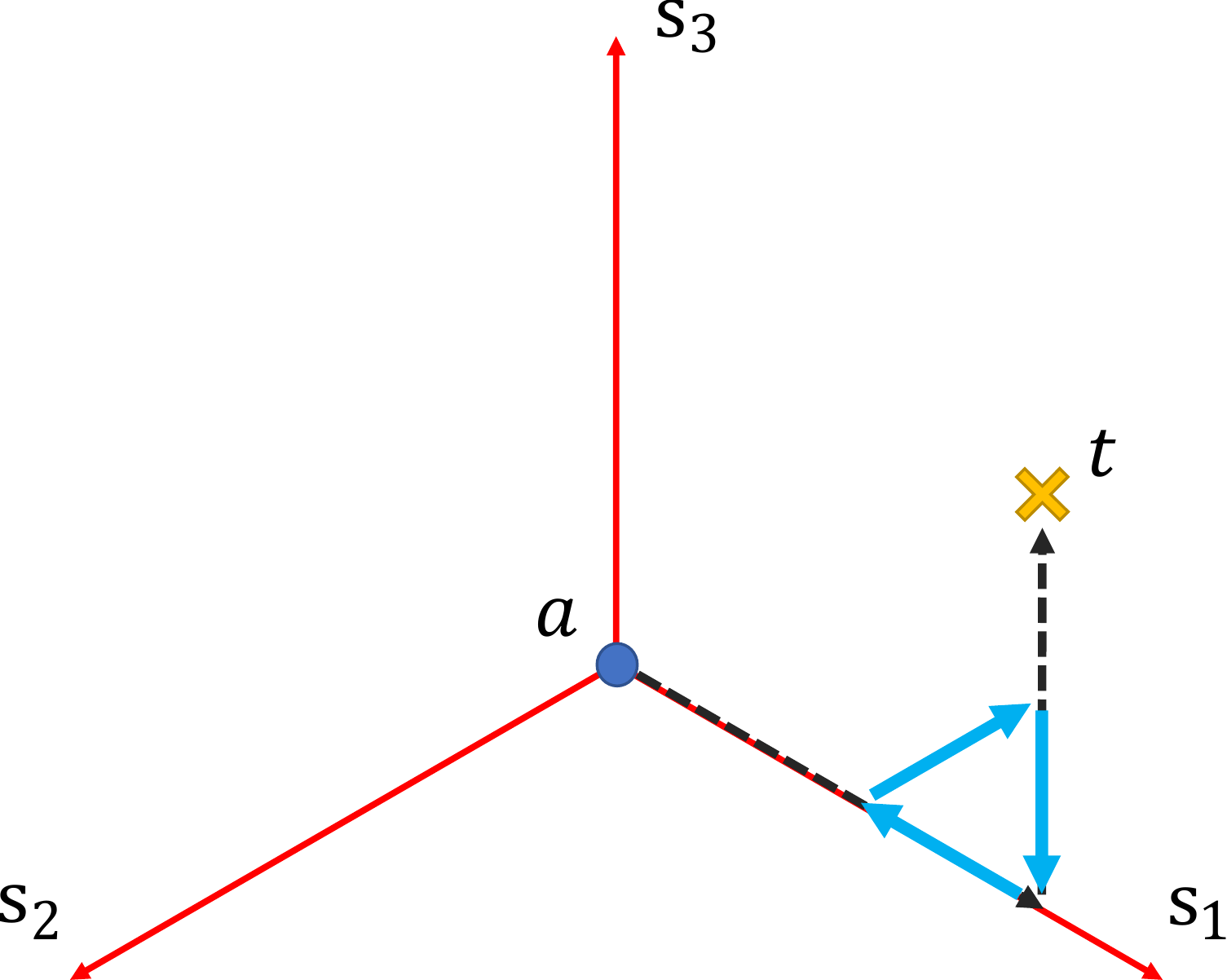}}\qquad
	\subfigure[\protect\url{}\label{fig:Cable_Geometry2_B}]%
	{\includegraphics[height=4cm]{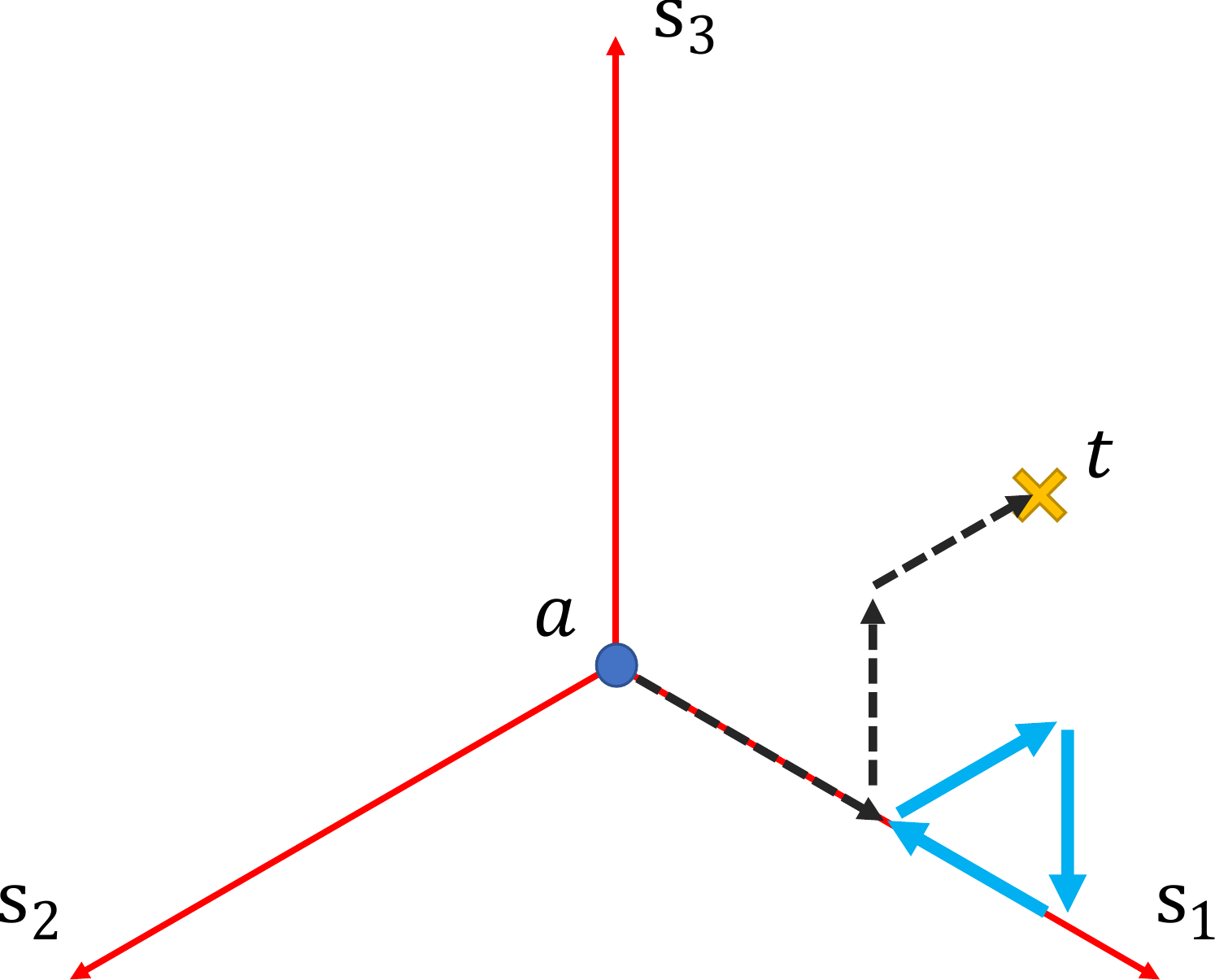}} \vspace{0.3cm}
	\caption{Alternative route to redistribute motor actuation. (a) Based on the inverse kinematics, going from $a$ to $t$ by only performing cable pulling would generate two positive commands on $s_1$ and $s_3$ (with $s_1 > s_3$), leaving $s_2$ to be zero. (b) Alternatively, a part $s_1$ can be transferred to the other two axes, resulting in two positive values on $s_1$ and $s_3$. The final actuation would be two motors pulling ($s_1$ and $s_3$) and one slightly releasing ($s_2$). The triangle represented in the figure has a side equal to the part of $s_1$ we intend to redistribute; this triangle is also equilateral, as the space is divided into three sectors having angles of $\frac{2}{3}\pi$ between each axis. Therefore, redistributing a value on opposite directions can be simply performed by addition or subtraction.}
	\label{fig:cable_geometry}
\end{figure}

The redistribution adds part of $t^m_{s_{max}}$ as cable release to all motors, which means that all positive commands should pull less and all negative commands should release more. We define $\epsilon \in [0,1]$ as a coefficient that removes part of $t^m_{s_{max}}$.
\begin{equation}\label{eq:redistribuition_sub}
    \forall j \in [1,3] \enskip : \enskip t^m_{s_j} \leftarrow t^m_{s_j} - t^m_{s_{max}} \cdot \epsilon
\end{equation}
Due to the geometry of the Motor space, this operation is trivial: any triangle having $a$ and $p$ as vertices and all sides parallel to the space axes is equilateral; therefore, adding or subtracting the same value to all the dimensions will provide a valid solution for~\eqref{eq:steeringSystem}, as shown in the example on Fig.~\ref{fig:cable_geometry}.

However, this procedure is not required when the direction of the robot remains the same in two consecutive iterations. 
Therefore, actuation redistribution is only performed when~\eqref{eq:redistribuition_condition} is true. Given the motor space coordinates of the actual end-effector position $a$, which represents the steering commands to go from the origin to the previous desired position, we compared it to a new position $t$. If both $t^m$ and $a^m$ share the same null coordinate, then the robot is moving in the same direction as in the previous iteration, and there is no need for cable release.
\begin{align}\label{eq:redistribuition_condition}
    \nexists i \in [1,3] \enskip : \enskip t^m_{s_i} = 0 \enskip \land \enskip a^m_{s_i} = 0
\end{align}
In our setup we set $\epsilon = 0.3$, as we found that higher values caused slack in the cables and lower values produced excessive wrinkling.

Finally, we transform the steering coordinates from distances to angles.
Sending angles rather than distances to the closed loop control is preferable, as the commands given to steer depend on the manipulator's length. 
Angles are calculated over the actual length of the robot $L_a$ as both $a$ and $p$ share this value. The error $\psi$ is then returned as output, where the steering commands are defined in $t^m$ and the eversion command is $L_d - L_a$.

\section{Closed-Loop Control}\label{sec:closed-loop-control}

Separate controllers are used for the eversion and steering DoFs because they have different dynamics. For eversion, we use feed-forward augmentation on a proportional controller. This augmentation is required to counter the passive growth that occurs from pressurizing the system. The proportional term, which is applied to an error of length in millimeters, proved sufficient for good tracking of growth, so we did not include an integral or derivative term. For steering, we use a scaled proportional-derivative controller on the error in rotation of the three motors; our gains are scaled proportionally with the length of the robot. 
This scaling is necessary because the steering dynamics are a function of the manipulator's length. We did not use an integral term because this is detrimental to telepresence during teleoperation, and we wanted to use the same low-level control strategy for all of the paradigms. 
All gains are hand tuned.